\documentclass{isprs}

\usepackage{graphicx}
\usepackage{epstopdf}
\usepackage{subcaption} % removed subfigure to avoid clashes
\usepackage{svg}
\usepackage[labelsep=period]{caption}
\usepackage{stfloats}

\usepackage{array}       
\usepackage{booktabs}    
\usepackage{multirow}    
\usepackage{tabularx}    
\usepackage{colortbl}    

\usepackage{amsmath,amssymb,amsfonts}
\usepackage{setspace}
\usepackage{geometry} 
\usepackage[british]{babel} 
\usepackage[hang]{footmisc}

\usepackage{cite}
\usepackage{textcomp}
\usepackage{xcolor,soul}
\usepackage{algorithm}
\usepackage{algpseudocode}
\usepackage{pifont}
\usepackage{multicol}
\usepackage{url}
\usepackage{lmodern}

\newcommand{\cmark}{\ding{51}}
\newcommand{\xmark}{\ding{55}}
\begin{document}

\title{MapAnything: \\ Evaluating Monocular Metric Depth Models for 3D Urban Asset Localization}
\date{}

\author{
  Miriam L. Carnot\textsuperscript{1}, 
  Jonas Kunze\textsuperscript{1}, 
  Erik Q. Fastermann\textsuperscript{2}, 
  Eric Peukert\textsuperscript{1}, 
  André Ludwig\textsuperscript{2,3}, 
  Bogdan Franczyk\textsuperscript{2,4}
  }
\address{
	\textsuperscript{1 } ScaDS.AI (University of Leipzig) \\
	\textsuperscript{2 } University of Leipzig \\
        \textsuperscript{3 } Kühne Logistics University \\
        \textsuperscript{4 } Wrocław University of Economics\\
}

% frame as a low-cost method for 3d data acquisition
\abstract{
    City administrations increasingly rely on comprehensive databases and digital twins of city assets, such as traffic signs and trees, as well as incidents such as graffiti or road damage, to maintain an effective overview of urban conditions. Digitization has increased the demand for continuously updated spatial datasets, yet current data acquisition and maintenance processes still involve considerable manual effort, posing significant scalability challenges. This paper introduces MapAnything, a systematic evaluation pipeline that automates the spatial mapping of urban objects and incidents from a single monocular image. By leveraging advanced Metric Depth Estimation models, MapAnything accurately calculates object geocoordinates, converting 2D image data into valuable 3D spatial information. The methodology integrates the estimated camera-to-object distance with geometric principles and known camera specifications. We present a detailed validation of the framework, comparing its distance-estimation accuracy against high-precision LiDAR point clouds in complex urban environments. Our evaluation provides a granular analysis of spatial performance across various distance intervals and semantic areas, such as roads and vegetation. Finally, we demonstrate the framework's practical efficacy through specific use cases, including mapping traffic signs and road pavement damage, and provide recommendations for its integration into automated urban inventory systems.
}

%To maintain an overview of urban conditions, city administrations manage databases of objects like traffic signs and trees, complete with their geocoordinates. Incidents such as graffiti or road damage are also relevant. As digitization increases, the need for more data and up-to-date databases grows, requiring significant manual effort.
%This paper introduces MapAnything, a module that automatically determines the geocoordinates of objects using individual images. Utilizing advanced Metric Depth Estimation models, MapAnything calculates geocoordinates based on the object's distance from the camera, geometric principles, and camera specifications.
%We detail and validate the module, providing recommendations for automating urban object and incident mapping. Our evaluation measures the accuracy of estimated distances against LiDAR point clouds in urban environments, analyzing performance across distance intervals and semantic areas like roads and vegetation. The module's effectiveness is demonstrated through use cases involving traffic signs and road damage.


\maketitle
\sloppy

\section{Introduction}
\label{sec:intro}

% motivation
Every city has countless infrastructure assets financed by public funds that help ensure people can live together efficiently, safely, and comfortably.  These include all traffic safety installations, such as traffic lights, signs, and street lamps, as well as recreational facilities such as benches, planting, bus stops, and bicycle stands.
City governments maintain spatial databases and digital twins for many of these assets, which record each asset's location, type, and, where applicable, condition. There are two fundamental challenges here:
1. the initial creation of the data record, and 2. keeping the database up to date. Many objects can be damaged or even be stolen over time. Thus, it is important to carry out regular checks. The opposite can also be true, for example, when new traffic signs are installed on a construction site. 
It may also be of interest to map incidents such as graffiti or road pavement damage. By equipping vehicles such as trams, taxis, and garbage trucks with cameras, the necessary images can be captured to provide a continuously updated, cost-effective spatial overview.

% problem statement
Current solutions for mapping objects have their pitfalls. Some use either additional aerial imagery or point clouds from LiDAR sensors \cite{wegner_cataloging_2016, branson_google_2018, boller_automated_2019}, which are not always available, as such recordings are expensive and therefore cannot be taken regularly. Other works merely use the camera's coordinates at the time of recording \cite{graffiti}, which is insufficient for accurate mapping. Yet other studies include knowledge of the object's size in the calculation \cite{campbell_detecting_2019}. This approach assumes that the objects searched for have the same size and that this size is known. Finally, some approaches analyze several consecutive images and then locate objects using triangular relationships \cite{hebbalaguppe2017telecom, cheng_crowd-sourced_2018, krylov_automatic_2018, krylov_object_2019}. This requirement that an object must be visible in several images is not always given. To address this, we introduce a highly scalable, low-cost method for 3D data acquisition. By developing a universal framework that extracts 3D spatial data from single 2D images, we eliminate the need for prior object knowledge or multiple overlapping viewpoints. This makes our approach particularly valuable for leveraging crowd-sourced imagery, which often lacks high-quality, corresponding LiDAR point clouds or consistent multi-view coverage.

% approach
To address this, our study is guided by the following research question: To what extent can state-of-the-art monocular metric depth estimation models be reliably utilized for the 3D geo-localization of urban assets, and how do they perform across varying distances and semantic backgrounds?

Our developed framework relies on these models to estimate the distance from the camera to each pixel in a single image. We examine four recently published state-of-the-art models: DepthAnything \cite{depth_anything}, DepthPro \cite{depth_pro}, Metric3D \cite{Metric3D-1, Metric3D-2}, and UniDepth \cite{unidepth}. To test their suitability for our practical urban application, we compare the generated depth images with projections of 3D LiDAR point clouds from the same scene, evaluating errors across semantic classes (such as roads, vegetation, and buildings) and distance ranges. With the estimated depth and the camera's position and orientation, the object's geo-coordinates are then extracted using angular deviations from the center point and triangulation. Finally, we evaluate two practical use cases: traffic signs and road pavement damage. In both cases, we assess localization accuracy, i.e., the distance between the predicted and true coordinates. For the traffic signs, we also determine how many of the annotated signs were found and compare the predictions to the city council's database.

% contributions
The main contributions of this paper include:
\begin{itemize}
    \item a systematic evaluation of a highly scalable, low-cost pipeline for extracting the geo-coordinates of urban assets from single 2D street-view images,
    \item a benchmarking of four SOTA monocular metric depth estimation models, assessing their distance-estimation accuracy against high-precision LiDAR point clouds across various urban semantic areas,
    \item a demonstration of practical utility through two real-world case studies (mapping traffic signs and road pavement damages), evaluating the reliability of the predicted geo-coordinates for updating digital twins.
\end{itemize}
%
%The City of Leipzig is our practical partner for this project and operates the data infrastructure for the image and point cloud data used in the paper. 
%
% chapters
%We start by analyzing existing methods to map objects based on images and give an overview of depth estimation in Section \ref{sec:rel_work}. We explain our methods in Section \ref{sec:methodology} and present results from our chosen depth models and two exemplary use cases in Section \ref{sec:experiments}. We finally discuss and conclude our findings in Section \ref{sec:discussion-conclusion}.
%
Code, annotations, and test data are available under: \url{https://github.com/miriamcarnot/MapAnything}.

\section{Related Work}
\label{sec:rel_work}
This Section discusses approaches to estimating the geo-location of urban objects and provides information on Monocular Depth Estimation, a core part of the MapAnything framework.

\subsection{Mapping Occurrences or Objects}

For the most part, data collection for urban objects in geographic information systems (GIS) and city models relies on high-precision, resource-intensive methods. City administrations typically use Mobile Mapping Systems (MMS) with LiDAR sensors, aerial photogrammetry, or manual terrestrial surveys to create and update precise digital twins of the city. While these approaches provide high-accuracy spatial data, they require expensive equipment, labor-intensive manual post-processing, and dedicated mapping campaigns. This significantly limits the frequency of data collection, making the continuous maintenance of an up-to-date city inventory a challenge both financially and in terms of scalability. To address these limitations, several studies have investigated the geo-referencing of objects using Street View imagery for urban applications.
The most basic method is to use the camera's geo-coordinates at the time of recording. For example, Novack et al. \cite{graffiti} used this approach for geo-referencing graffiti on building façades.
Another possible solution is to include aerial images that provide additional information. Boller et al. \cite{boller_automated_2019} use high-quality orthophotos to develop a mapping method for panorama images. They present a formula for calculating the geo-coordinates that also uses the distance between the object and the camera, but they do not reveal its origin. Their evaluation is purely qualitative. Another study that used street-level and aerial imagery tested their method on trees \cite{wegner_cataloging_2016}. In a follow-up study, they also classified tree species and tracked and classified changes \cite{branson_google_2018}. 

A third valid approach is combining the information from multiple images. These approaches assume that objects can be seen across multiple images and are therefore not suitable for all types of data collection. 
One example of this line of research is the work by Hebbalaguppe et al. \cite{hebbalaguppe2017telecom}. They detect telecommunication inventory in GSV image pairs of consecutive locations to perform triangulation and estimate coordinates.
One research group also used a depth estimation model to retrieve the distance to the object. They estimate depth and then fuse distance information across multiple images using Markov Random Fields for Triangulation. In their first work \cite{krylov_automatic_2018}, they detect traffic lights and telegraph poles in Google Street-View (GSV) images while they look into crowd-sourced image data (Mapillary) in their second study \cite{krylov_object_2019}.
If a large number of images are available, one may also use photogrammetry. Cheng et al. \cite{cheng_crowd-sourced_2018} retrieve many crowd-sourced images of one building and use them to generate a point cloud. The location and extension of the building can then be determined from the created point cloud. 

Some mapping approaches are object-specific, such as the one described by Campbell et al. \cite{campbell_detecting_2019}. They geo-reference traffic signs in GSV images using information about their dimensions. They determine the distance between the sign and the camera by calculating the ratio of the sign's real-world size to its image-space size. Then they calculate the angular offsets relative to the object in the image to determine its coordinates.
Another study by Vishnani et al. \cite{vishnani_manhole_2020} uses a simple regression model and the Haversine formula to map manholes detected in GSV imagery. However, it is not further explained or evaluated.
In our work, we aim to determine the geographic coordinates of an urban object without any additional records or information about it, though we acknowledge that relying solely on single-image data introduces inherent limitations in absolute accuracy compared to multi-modal approaches.

Similar localization challenges arise in autonomous driving and robotics, yet their primary objective is to detect 3D objects from 2D images \cite{monogrnet, Li2023Keypoint3DKA}. Because their focus is on navigation and obstacle avoidance, these networks are designed to extract precise 3D bounding boxes and 6-DoF poses for dynamic objects (e.g., vehicles). While many of these models incorporate depth estimation, they prioritize extracting precise 3D bounding boxes and object orientation. In contrast, finding the geo-coordinates of static urban objects does not require finding the bounding box and its orientation. Furthermore, to achieve high reliability, the most robust methods in these domains typically rely on multi-modal or multi-view data rather than a single image, utilizing LiDAR \cite{zhang2024safdnet}, Radar \cite{lin2024rcbevdet}, RGB-D cameras \cite{frustrumpointnets}, or multiple images taken simultaneously \cite{li2023bevdepth} or in sequence.

\subsection{Monocular Depth Estimation}
Depth estimation aims to infer the third dimension of a recorded scene, estimating the distance of each pixel from the camera. Monocular depth estimation performs this task using a single image, without stereo or multi-view information \cite{Masoumian2022}.
The task can be categorized into relative and metric depth estimation. Relative depth estimation focuses on capturing depth relationships between objects in an image, ensuring correct ordering rather than absolute scale. Metric depth estimation, in contrast, aims to predict absolute distances between objects and the camera, requiring scale awareness through additional priors or estimations \cite{zhu2024scaledepth}. 
Early monocular depth estimation models relied on supervised learning, using single-dataset training with consistent camera parameters and recording conditions (e.g., focal length, field of view, and lighting) \cite{Eigen2014, Garg2016, Godard2016}. These models often lacked generalization across diverse environments. Later, models such as MiDaS \cite{midas_model} introduced multi-dataset training, significantly improving cross-dataset performance. 
Recently, zero-shot monocular depth models have demonstrated robust generalization across diverse images, handling varying domains without fine-tuning \cite{zoedepth, depth_pro}. 
%Depth Anything \cite{depth_anything} also exhibits strong zero-shot capabilities through its training on large-scale datasets.
Within the last few years, several new models beating previous results were introduced to estimate metric depth: DepthAnything \cite{depth_anything}, Depth Pro \cite{depth_pro}, UniDepth \cite{unidepth}, and Metric3D \cite{Metric3D-1}\cite{Metric3D-2}, among others. These models use additional constraints, such as learned geometric priors or scale normalization, to achieve better predictions of absolute depth values. So far, these models have only been evaluated on known benchmarks, not yet for concrete applicability to street-view images in practical urban use cases.
\section{Methodology}
\label{sec:methodology}

In this Section, we present the three key steps to map any urban object or occurrence with a single street-view image:
\begin{enumerate}
    \item \textit{Detect/Segment the desired object} in the image.
    \item Apply the \textit{MapAnything Framework}:
        \begin{enumerate}
            \item{Estimate the metric depth for every pixel.}
            \item{Calculate each object's coordinate based on the predicted depth at this position and its angular offset from the image's center.}
       \end{enumerate}
    \item \textit{Remove duplicates} of objects that were recognized in multiple images and/or match with database entries.
\end{enumerate}

\begin{figure*}[th]
    \centering
    % First row with one image
    \begin{subfigure}[b]{0.3\textwidth}
        \centering
        \includegraphics[width=\linewidth]{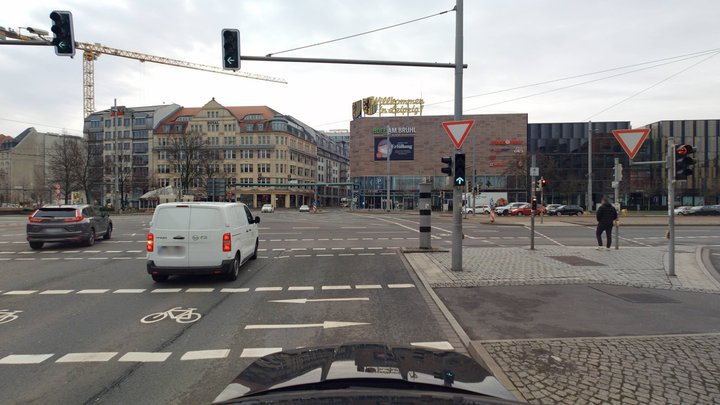}
        \caption{Original Image}
    \end{subfigure}  
    \begin{subfigure}[b]{0.32\textwidth}
        \centering
        \includegraphics[width=\linewidth]{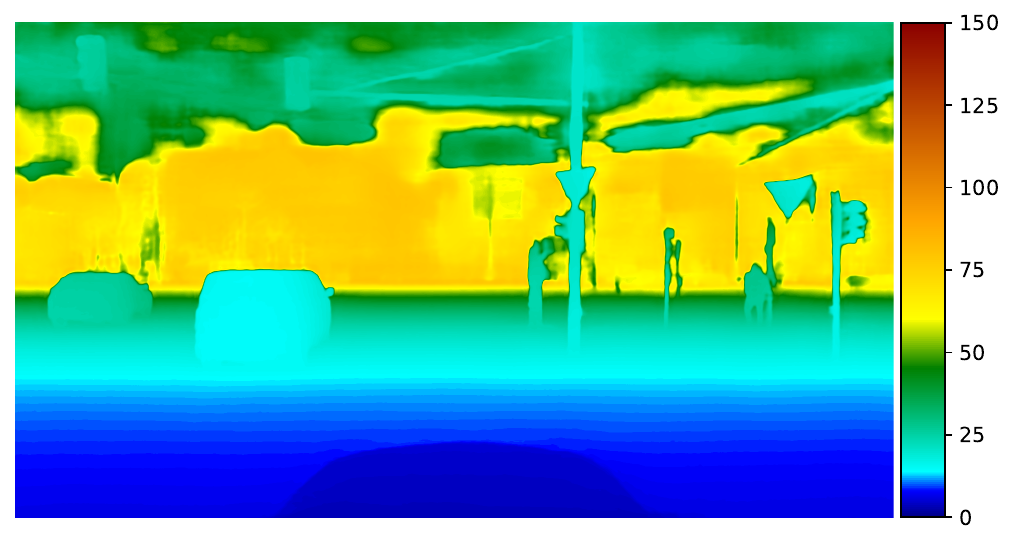}
        \caption{DepthAnything}
    \end{subfigure}
    \begin{subfigure}[b]{0.32\textwidth}
        \centering
        \includegraphics[width=\linewidth]{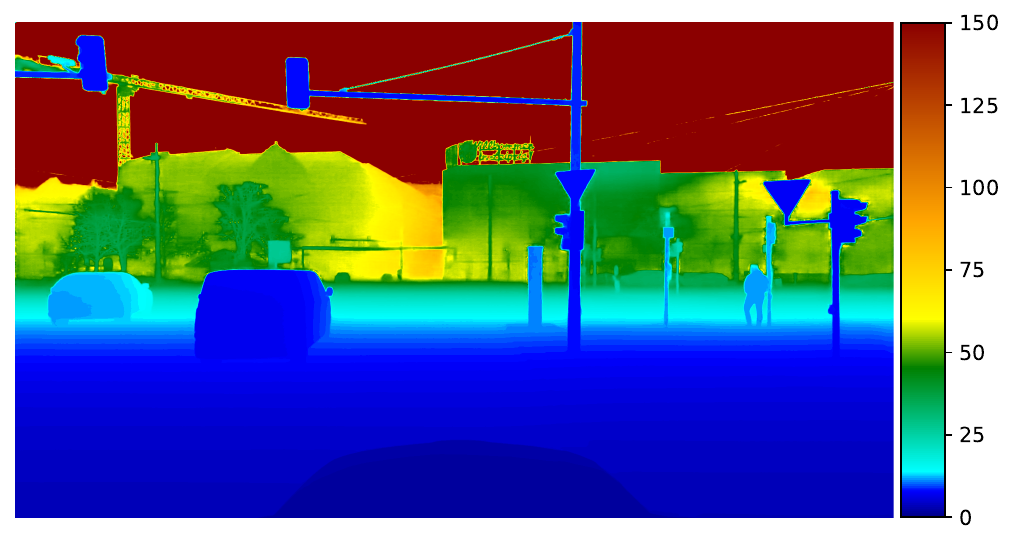}
        \caption{DepthPro}
    \end{subfigure} \\
    \vspace{2mm}
    \begin{subfigure}[b]{0.32\textwidth}
        \centering
        \includegraphics[width=\linewidth]{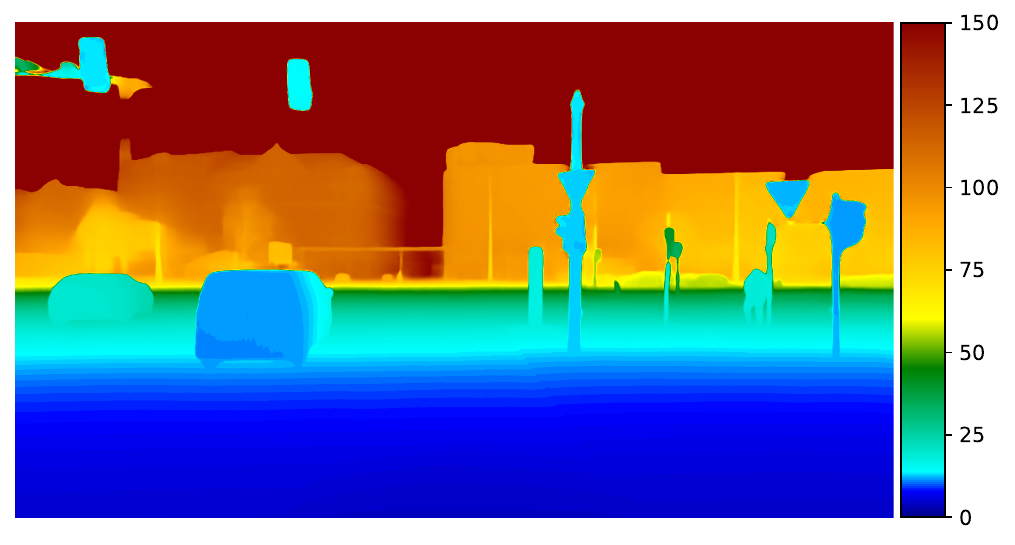}
        \caption{UniDepth}
    \end{subfigure}
    \begin{subfigure}[b]{0.32\textwidth}
        \centering
        \includegraphics[width=\linewidth]{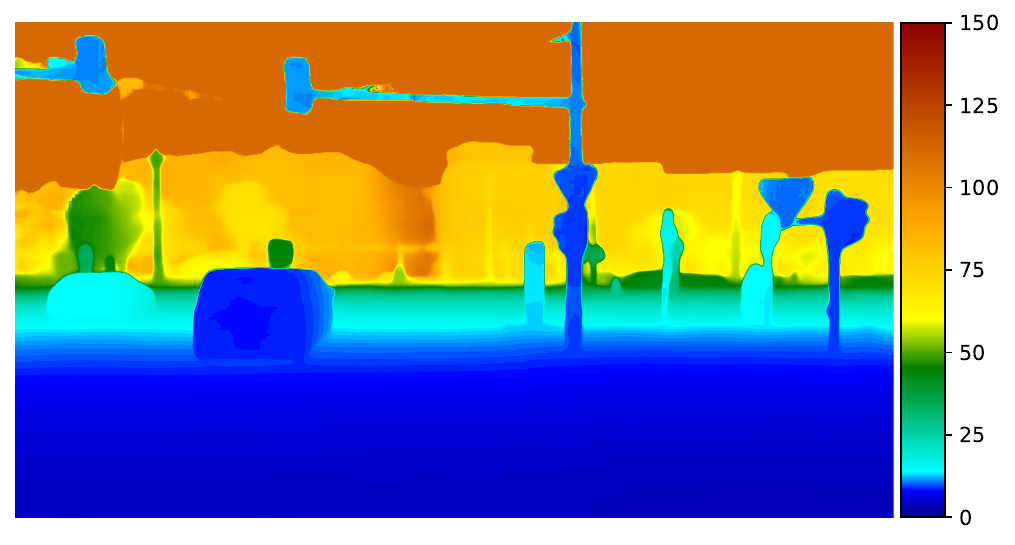}
        \caption{Metric3D}
    \end{subfigure}

    \caption{Comparative visualization of four SOTA Monocular Metric Depth Estimation models. Scales are in meters. For better visibility, we clip the estimations at 150m. Note how DepthPro (c) maintains sharper edges on the poles compared to the over-smoothed output of Metric3D (e), which impacts the precision of the resulting geocoordinates.}
    \label{fig:depth_images}
\end{figure*}

\subsection{Detect/Segment the Object in the Image}
To map an object, the system needs to know where in the image it appears. These image areas can be annotated manually or predicted by a model. There are freely available models for detecting or segmenting various objects on platforms such as HuggingFace. Of course, it is also possible to train a model from scratch or fine-tune it on specific data.
Among the valuable datasets containing large quantities of annotated urban images are Mapillary Vistas \cite{mapillary}, Cityscapes \cite{cityscapes}, and A2D2 \cite{a2d2}. 
For our experimental evaluation (Section \ref{sec:experiments}), we use a U-Net trained on the A2D2 dataset for traffic-sign segmentation and a Segformer trained on the Cityscapes dataset for general semantic segmentation.

\subsection{The MapAnything Framework}
To calculate the geographic coordinates of an object, two key pieces of information are required: the camera's geo-coordinates at the time of recording and the recording direction, i.e., the deviation from north (yaw). All other camera information is not strictly necessary, as it can be estimated.
If unknown, the camera's focal length can be estimated directly from the image utilizing integrated modules within contemporary depth models, such as DepthPro. And the camera's pitch can be estimated based on visual features in the image (such as the horizon or buildings). 
In our approach, we use a depth estimation model to predict a dense metric depth map, assigning a depth value (in meters) to every pixel. We then compute the overall distance to the target object using either the depth at the center pixel of its 2D bounding box or a weighted average of the depths within its segmentation mask. For larger objects or occurrences, a coordinate range can be determined between the outermost points, e.g., for a widespread graffiti painting.
Four recent Monocular Metric Depth Estimation models and their variations will be evaluated in Section \ref{sec:experiments}.

Applying the standard pinhole camera model, we set up the following equations.
First, we normalize the pixel coordinates using the focal lengths \( f_x \) and \( f_y \):
\begin{equation}
x_{norm} = \frac{x - W/2}{f_x}, \quad y_{norm} = \frac{y - H/2}{f_y}
\end{equation}
Next, we determine the angular offsets \(\alpha_x\) and \(\alpha_y\) relative to the camera’s optical axis, effectively converting the image-plane coordinates into real-world angles:
\begin{equation}
\alpha_x = \tan^{-1}(x_{norm}), \quad \alpha_y = \tan^{-1}(y_{norm})
\end{equation}
We adjust the camera's yaw and pitch angles by adding the calculated angles:
\begin{equation}
\theta_{yaw, eff} = \theta_{yaw} + \alpha_x, \theta_{pitch, eff} = \theta_{pitch} + \alpha_y
\end{equation}
Using the predicted metric depth $d$ of the object, we first compute the horizontal distance $d_{horizontal}$ on the ground plane. We can then calculate the spatial displacements $\Delta x$ (eastward distance) and $\Delta y$ (northward distance) relative to the camera:
\begin{equation}
d_{horizontal} = d \cdot \cos(\theta_{pitch, eff})
\end{equation}
\begin{equation}
\Delta x = d_{horizontal} \cdot \cos(\theta_{yaw, eff})
\end{equation}
\begin{equation}
\Delta y = d_{horizontal} \cdot \sin(\theta_{yaw, eff})
\end{equation}
Finally, we convert the displacements into changes in latitude and longitude using the Earth's radius \( R \) and add them to the camera's coordinates:
\begin{equation}
\Delta \text{lat} = \frac{\Delta y}{R}, \quad \Delta \text{lon} = \frac{\Delta x}{R \cos(\text{lat}_{camera})}
\end{equation}
\begin{equation}
\text{lat}_{object} = \text{lat}_{camera} + \left(\frac{180}{\pi}\right) \cdot \Delta \text{lat}
\end{equation}
\begin{equation}
\text{lon}_{object} = \text{lon}_{camera} + \left(\frac{180}{\pi}\right) \cdot \Delta \text{lon}
\end{equation}
Based on these calculations, there is a linear relationship between the deviation in the depth determination and the error of the calculated coordinate.

\subsection{Remove Duplicates or match with Database}
To identify duplicates resulting from overlapping images, we group predictions that share the same object class and fall within a specified geographic radius. The radius size that represents the best trade-off between correct matches and error rate should be determined for each object type. A basic approach to determining this radius is to analyze the minimum distance between two distinct ground-truth objects of the same class. When duplicates are found within this range, we take the center of the predicted coordinates as the final prediction. 

While our experimental prototype uses a straightforward radius search to assess localization accuracy, we acknowledge that this $O(n^2)$ brute-force comparison does not scale well. For practical, city-scale deployments involving massive datasets, integrating spatial indexing structures (such as R-trees) into the spatial database is essential for efficiently processing spatial queries.
Furthermore, to match our finalized predictions with existing municipal database entries, we use the calculated angular offsets (added to the recording direction) to find entries of the matching class within a reasonable distance, accounting for slight angle deviations.
\section{Experiments}
\label{sec:experiments}
We first analyze the extent to which the predicted depth deviates from the actual distance measured by a LiDAR sensor. For a comprehensive evaluation, we test various models, distance range, and semantic segments in the images.
Then, we demonstrate the proposed pipeline in two use cases: traffic signs and road pavement damage, as these issues are common on most roads.

\subsection{Evaluation of the Depth Models}

We evaluate four models that were recently published: DepthAnything \cite{depth_anything}, DepthPro \cite{depth_pro}, UniDepth \cite{unidepth}, and Metric3D \cite{Metric3D-1}. 
DepthAnything offers three different model sizes, does not require any camera intrinsic information, and was trained on the synthetic Virtual KITTI dataset \cite{Cabon2020VirtualK2}.
DepthPro was trained on a combination of real-world and synthetic datasets to improve its generalization. It uses the focal length as an input parameter for scaling, but can also operate without this parameter by estimating it. We test both versions.
UniDepth (Version 2) also does not rely on external information and can estimate the camera's intrinsic parameters. 
%Even though it is possible to also provide the true camera intrinsic to the model, we obtain the same results with or without it. We found a problem in its implementation, which causes the model not to use the given parameters. We thus only use the model without the intrinsic for our study. 
It has a ViT backbone and is also available in three sizes.
Metric3D was trained on 18 datasets, including over 16 million images from thousands of different camera types. It always requires the camera intrinsic, and there are versions with a ViT backbone and another with a ConvNeXt backbone. 
\begin{table*}[t]
    \centering
    \scriptsize
    \setlength{\tabcolsep}{3pt}
    \begin{tabular}{@{} l c c c | c c c c c c c c | c c c c c c c c | c @{}}
        \toprule
        Model & cam & \multicolumn{2}{c}{total} & \multicolumn{2}{c}{\textless{}5m} & \multicolumn{2}{c}{5--10m} & \multicolumn{2}{c}{10--20m} & \multicolumn{2}{c}{\textgreater{}20m} & \multicolumn{2}{c}{flat} & \multicolumn{2}{c}{constr.} & \multicolumn{2}{c}{object} & \multicolumn{2}{c}{nature} & time \\
        \cmidrule(lr){3-4} \cmidrule(lr){5-6} \cmidrule(lr){7-8} \cmidrule(lr){9-10} \cmidrule(lr){11-12} \cmidrule(lr){13-14} \cmidrule(lr){15-16} \cmidrule(lr){17-18} \cmidrule(lr){19-20}
        & intr. & M & A & M & A & M & A & M & A & M & A & M & A & M & A & M & A & M & A & (s) \\
        \midrule
        DAny-S    & \xmark & 8.7 & 0.500 & \underline{1.3} & 0.275 & 3.7 & 0.517 & 9.4 & 0.655 & 13.9 & 0.398 & 4.4 & 0.465 & 10.9 & 0.452 & 12.4 & 0.649 & \underline{14.1} & \underline{0.562} & \underline{0.9} \\
        DAny-B    & \xmark & 9.0 & 0.524 & 1.5 & 0.327 & 3.9 & 0.545 & 9.7 & 0.676 & 14.4 & 0.420 & 4.7 & 0.496 & 10.9 & 0.463 & 12.1 & 0.643 & 14.7 & 0.591 & 2.3 \\
        DAny-L    & \xmark & 9.4 & 0.546 & 1.6 & 0.350 & 4.1 & 0.568 & 10.0 & 0.691 & 15.1 & 0.447 & 5.0 & 0.527 & 11.5 & 0.488 & 12.0 & 0.632 & 14.7 & 0.595 & 6.9 \\
        DPro-1    & \xmark & 13.7 & 0.697 & 2.2 & 0.484 & 3.2 & 0.460 & 13.8 & 0.968 & 26.9 & 0.805 & 2.4 & 0.283 & 9.9 & 0.309 & \underline{10.5} & \underline{0.474} & 24.4 & 0.936 & 7.1 \\
        DPro-2    & \cmark & 16.3 & 0.834 & 2.9 & 0.630 & 4.3 & 0.628 & 15.5 & 1.081 & 31.6 & 0.924 & 4.1 & 0.469 & 14.9 & 0.487 & 13.6 & 0.598 & 25.6 & 0.938 & 7.2 \\
        UniD-S    & \xmark & 6.9 & 0.331 & \textbf{1.2} & \textbf{0.253} & 2.0 & 0.271 & 6.1 & 0.424 & 13.3 & 0.353 & \underline{2.2} & 0.232 & 7.5 & 0.244 & 14.5 & 0.718 & 16.2 & 0.582 & \textbf{0.4} \\
        UniD-B    & \xmark & \underline{5.7} & \underline{0.267} & \underline{1.3} & 0.288 & \textbf{1.4} & \textbf{0.198} & \underline{5.0} & \underline{0.344} & \textbf{11.0} & \underline{0.283} & \textbf{1.5} & \underline{0.166} & \textbf{5.5} & \textbf{0.176} & 11.4 & 0.562 & 14.6 & 0.516 & \underline{0.9} \\
        UniD-L    & \xmark & 6.5 & 0.298 & \textbf{1.2} & \underline{0.263} & \underline{1.5} & \underline{0.199} & 5.9 & 0.410 & 12.5 & 0.329 & \textbf{1.5} & \textbf{0.154} & \underline{5.8} & \underline{0.181} & 15.1 & 0.781 & 17.6 & 0.644 & 2.5 \\
        M3D-V   & \cmark & \textbf{5.6} & \textbf{0.251} & 1.7 & 0.360 & \underline{1.5} & 0.219 & \textbf{3.5} & \textbf{0.237} & \underline{11.4} & \textbf{0.271} & \textbf{1.5} & 0.202 & 6.4  & 0.213 & \textbf{6.5}  & \textbf{0.278} & \textbf{13.0} & \textbf{0.350} & 3.2 \\
        M3D-C   & \cmark & 16.7 & 0.767 & 2.4 & 0.511 & 4.2 & 0.598 & 11.8 & 0.807 & 34.3 & 0.927 & 5.6 & 0.601 & 27.0 & 0.915 & 19.3 & 0.864 & 25.6 & 0.829 & 1.8 \\
        \bottomrule
    \end{tabular}
    \caption{Mean Absolute Error (M) and Absolute Relative Error (A) evaluated across depth ranges and semantic groups, alongside inference time per image. Model abbreviations: DAny (DepthAnything), DPro (DepthPro), UniD (UniDepth-v2), M3D-V (Metric3D-ViT), M3D-C (Metric3D-ConvNext). S/B/L denote Small, Base, and Large variants.}
    \label{tab:eval_depth} 
\end{table*}
\begin{figure*}[t]
    \centering

    \begin{subfigure}[b]{0.4\textwidth}
        \centering
        \includegraphics[width=\linewidth]{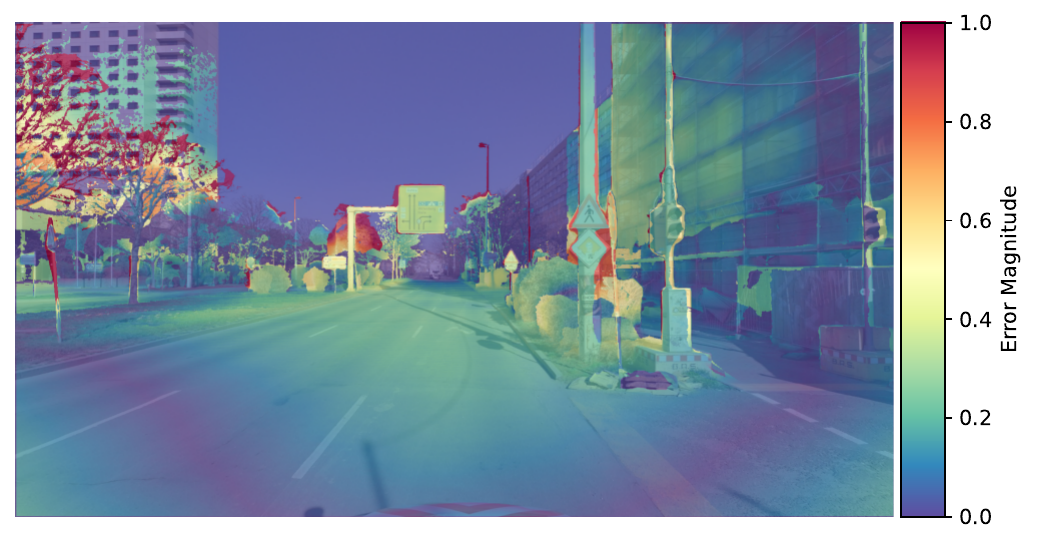}
        \caption{Absolute Relative Error}
    \end{subfigure}
    \hspace{5mm}
    \begin{subfigure}[b]{0.4\textwidth}
        \centering
        \includegraphics[width=\linewidth]{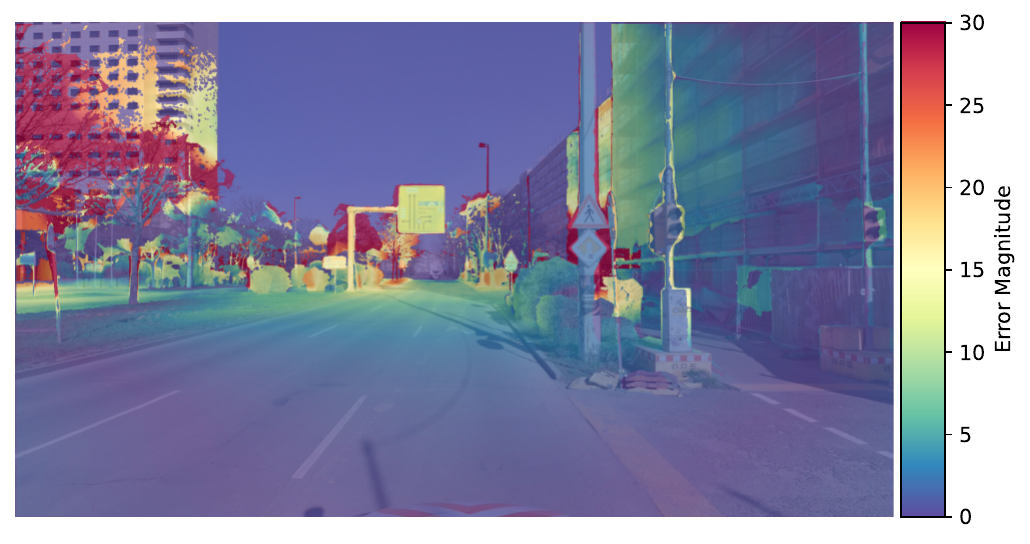}
        \caption{Mean Absolute Error}
    \end{subfigure}
    
    \vspace{2mm}
    
    \begin{subfigure}[b]{0.37\textwidth}
        \centering
        \includegraphics[width=\linewidth]{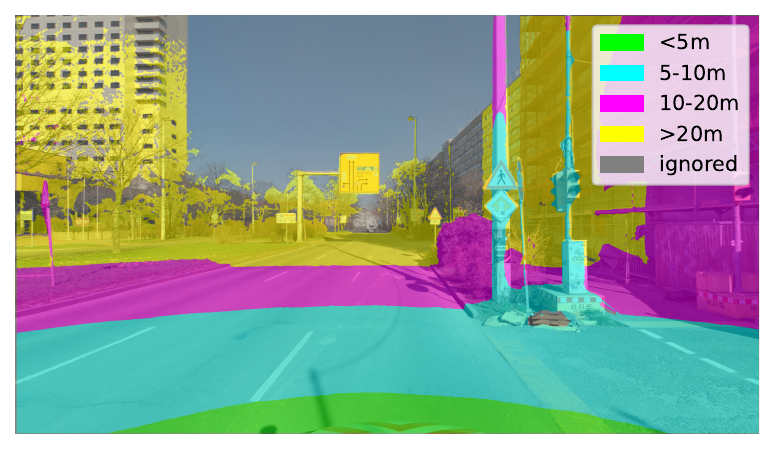}
        \caption{Distance Ranges}
    \end{subfigure}
    \hspace{7mm}
    \begin{subfigure}[b]{0.37\textwidth}
        \centering
        \includegraphics[width=\linewidth]{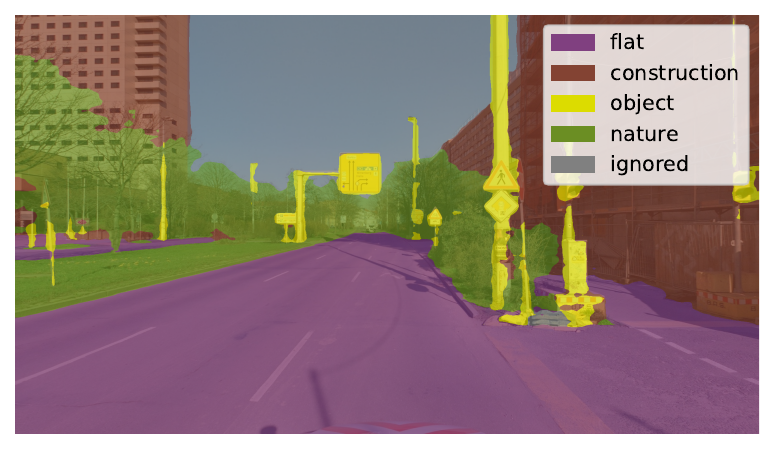}
        \caption{Semantic Groups}
    \end{subfigure}
    
    \caption{Visualization of depth estimation and segmentation metrics. Top: Errors produced by UniDepth. Bottom: Evaluation groupings from Segformer predictions (vehicles, people, and sky are ignored).}
    
    \label{fig:error_and_segmentation_vis}
\end{figure*}
\begin{table*}[t]
    \centering
    \scriptsize
    \begin{minipage}[t]{0.51\linewidth}
        \centering
        \textbf{(a) Traffic Signs Geo-Localization}\par\vspace{4pt}
        \footnotesize
        \setlength{\tabcolsep}{3pt}
        \begin{tabular}{@{} l c c c c c c c @{}}
            \toprule
            & \textbf{cam} & \multicolumn{2}{c}{\textbf{signs}} & \multicolumn{4}{c}{\textbf{avg dist (m)}}\\
            \cmidrule(lr){3-4} \cmidrule(l){5-8}
            & \textbf{intr.} & all & temp. & total & $<$10 & 10--20 & $>$20 \\ 
            \midrule
            \textbf{Database} & -- & 73\% (190) & 9\% (2) & 2.3 & -- & -- & -- \\
            \midrule
            \textbf{Cyclomedia} & & \textit{of 258} & \textit{of 23} & & & & \\
            DAny-B & \xmark & 52\% (134) & 26\% (6) & 6.3 & 4.6 & 6.7 & 6.3 \\
            DPro & \cmark & \underline{86\% (221)} & \textbf{100\%} (23) & \underline{4.8} & \underline{3.6} & \underline{4.3} & \textbf{5.7} \\
            UniD-B & \xmark & \textbf{87\% (225)} & \textbf{100\%} (23) & \textbf{4.4} & \textbf{3.0} & \textbf{3.7} & \textbf{5.7} \\
            M3D-V & \cmark & 76\% (195) & \textbf{100\%} (23) & 7.3 & 5.9 & 7.5 & 8.8 \\
            \midrule
            \textbf{Mapillary} & & \textit{of 86} & \textit{of 9} & & & & \\
            DAany-B & \xmark & 28\% (24) & 11\% (1) & 6.2 & 6.4 & 6.1 & \textbf{6.2} \\
            DPro & \cmark & \underline{52\% (45)} & \textbf{44\%} (4) & \underline{6.0} & 5.9 & \textbf{5.3} & 7.1 \\
            UniD-B & \xmark & \textbf{59\% (51)} & 22\% (2) & \textbf{5.7} & \textbf{5.1} & \underline{5.4} & \underline{6.3} \\
            M3D-V & \cmark & 50\% (43) & \underline{33\%} (3) & 6.6 & \underline{5.4} & 7.5 & 8.8 \\
            \bottomrule
        \end{tabular}
    \end{minipage}% 
    \hfill%
    \begin{minipage}[t]{0.47\linewidth}
        \centering
        
        \textbf{(b) Ablation Study: Coordinate Error}\par\vspace{4pt}
        \footnotesize
        \begin{tabular}{@{} l c | c c @{}}
            \toprule
            & & \multicolumn{2}{c}{\textbf{Depth}} \\
            & & Annotation & Estimation \\
            \midrule
            \multirow{2}{*}{\textbf{Segm.}} & Annotation & 0 & 2.17 \\
            & Prediction & 0.06 & 2.21 \\
            \bottomrule
        \end{tabular}
        
        \vspace{12pt} 
        
        \textbf{(c) Road Damages Geo-Localization}\par\vspace{4pt}
        \footnotesize
        \setlength{\tabcolsep}{4pt}
        \begin{tabular}{@{} l c c c c c c @{}}
            \toprule
            & \textbf{cam} & \multicolumn{5}{c}{\textbf{avg. distance (m)}}\\
            \cmidrule(l){3-7}
            & \textbf{intr.} & total & 2--4 & 4--6 & 6--8 & 8--10 \\
            \midrule
            DAny-B  & \xmark & \textbf{2.34} & 2.9 & 3.2 & \textbf{2.1} & \textbf{2.2} \\
            DPro    & \cmark & 3.86 & 2.5 & 3.0 & 4.2 & 4.6 \\
            UniD-B  & \xmark & \underline{2.80} & \textbf{1.5} & \textbf{2.2} & \underline{2.5} & \underline{2.8} \\
            M3D-V & \cmark & 4.42 & \underline{2.0} & \underline{2.5} & 3.4 & 3.8 \\
            \bottomrule
        \end{tabular}
    \end{minipage}
    
    \vspace{1pt}
    \caption{Case Studies and ablation. \textbf{(a)} Annotated signs present in the database and/or predicted by our system, alongside average distance to the true geo-location. \textbf{(b)} Mean coordinate error (in meters) introduced across 20 sample images depending on segmentation and depth sources. \textbf{(c)} Coordinate error for road damages across distance intervals. For model abbreviations, see Table~\ref{tab:eval_depth}.}
    \label{tab:case_studies}
\end{table*}
Figure \ref{fig:depth_images} shows predicted depth maps from the original image in (a). 
It is important to note that sky pixels are usually mapped to the highest possible value. The Metric3D model only predicts depths up to 100 meters. Thus, the sky appears orange (its maximum) instead of red.
DepthAnything (b) is the only model that struggles with the sky and the background, but it produces good results for objects close to the camera. The depth map of Metric3D (e) appears blurred, making it difficult to recognize contours, such as the pedestrian on the right. Depth Pro (c) produces very fine contours, even for the pipe of the traffic light, which UniDepth (d) does not find at all. Overall, UniDepth and Metric3D seem to agree on how far objects are away from the camera. For example, both estimate the background buildings to be about 80 meters away, whereas DepthPro predicts about 50 meters. 

To evaluate these models on our data, we match 20 sample images with LiDAR point clouds recorded at the same position. 
To quantify the deviation between the measured and predicted distances, we calculate the mean absolute error (MAE) in meters (how far off is the estimation) and the absolute relative error (ARE):
\begin{equation}
    ARE = \frac{\left| depth_{pred} - depth_{true} \right|}{depth_{true}}
\end{equation}
The ARE normalizes the mean deviation. Large deviations in the background, therefore, do not influence the metric more than small deviations near the camera.
We do not calculate errors for pixels in the image that were segmented as one of the person or vehicle classes, as moving objects were removed from the point clouds, which would otherwise lead to errors in the calculations.
In Figure \ref{fig:error_and_segmentation_vis}, we visualize the ARE (a) and MAE (b) for an example image using the UniDepth model. We can see that the MAE gradually increases with distance from the camera, whereas the normalized ARE remains constant.
In addition to the total deviation, we examine the deviation at specific distance intervals: less than 5m, 5-10m, 10-20m, and over 20m. 
We also analyze different semantic areas in the image by applying a Segformer model trained on the Cityscapes dataset \footnote{https://huggingface.co/nvidia/segformer-b5-finetuned-cityscapes-1024-1024}. We group the classes into four semantic groups: flat (road, sidewalk, parking, rail track), construction (building, wall, fence, guard rail, bridge, tunnel), object (pole, pole group, traffic sign, traffic light), and nature (vegetation, terrain) - leaving out vehicles, people, and sky classes. 
Figure \ref{fig:error_and_segmentation_vis} also shows the defined distance ranges (c) and semantic groups (d) for an example image.
We also measure the inference time per image on an RTX 4090 GPU. 

The results for the different depth ranges are shown in the left half of Table \ref{tab:eval_depth}. Overall, and for each distance range, either the UniDepth or the Metric3D model with the ViT backbone produces the results closest to the point clouds. Metric3D has slightly lower overall absolute (MAE) and normalized (ARE) errors. UniDepth achieves better results for smaller distance ranges, while Metric3D is better for objects farther from the camera. 

The results for the semantic groups, shown on the right side of Table \ref{tab:eval_depth}, show a similar overall picture, with UniDepth and Metric3D-ViT achieving the best scores. UniDepth works better for large man-made structures represented by the flat and construction groups, while Metric3D performs better for the object and nature groups. All models struggle with the nature group. They might not catch the details of single branches of a tree.
For the models, it is easiest to estimate depth in flat structures. Roads, for example, are very uniform across different datasets. 

Regarding inference time, the small UniDepth model is the fastest, followed by the small DepthAnything and the basic UniDepth models. DepthPro takes the longest, even when the focal length is provided. Also, the large version of DepthAnything takes considerable time. 
DepthPro includes a module to estimate the focal length of the source image. While the camera's true focal length is 960 pixels, the average predicted focal length is 1433 pixels (ranging from 1079 to 1597). Nevertheless, the model that uses the predicted focal length instead of the ground truth achieves better results. Overall, we do not observe a clear advantage for models that use camera intrinsics as input, such as Metric3D, compared to models that do not use them, such as UniDepth.

\subsection{Case Study 1: Traffic Signs}
We annotated an urban region encompassing 4.74 km of road with 261 traffic signs (23 of which were temporarily mounted). We evaluated our approach using high-resolution imagery from Cyclomedia (786 images from early 2024), commissioned specifically for street recording, as well as crowdsourced images from Mapillary (231 images from October 2023 to April 2024). Based on OpenStreetMap data, the Cyclomedia images provide 100\% street coverage, whereas the Mapillary images cover only 28\% of streets during the corresponding timeframe. By analyzing the driving direction, we determined the theoretical visibility of the signs: 258 signs should be visible in the Cyclomedia data (including all temporary signs), compared to 86 in the Mapillary data (including 23 temporary signs). A key challenge with crowd-sourced data is GPS inaccuracy. The recorded camera coordinates differ by several meters from their actual positions, resulting in significant localization errors for the objects. For twelve manually measured images, we found a mean deviation of 5.5 meters.

Table \ref{tab:case_studies} (a) details the number of theoretically visible signs that were successfully detected in both image sources. We also compare the existing database entries against these manual annotations. %Area maps detailing camera positions, viewing directions, and sign locations can be found in the supplementary material.

Signs are segmented with a U-Net trained on the A2D2 dataset, then classified with a Vision Transformer (ViT) into the classes of the German Road Traffic Regulations, which are also used in the city's database. The classification model achieved 98.6\% test accuracy. We performed an ablation study with 20 images, analyzing how much of the coordinate offset was introduced by the depth estimation vs. segmentation. Results in Table \ref{tab:case_studies} (b) show that the segmentation step contributes only minimally to the overall error. 

To establish the de-duplication radius, we determined the minimum distance between two signs of the same class. The closest instance was 2.69 meters for two parking signs. Consequently, we adopted a three-meter radius to filter duplicate predictions generated from distinct images in the same area. To match our predictions with the annotations, we assigned each predicted traffic sign to the nearest marked traffic sign of the same class, using a maximum distance threshold of ten meters.

Our results demonstrate that, when applied to high-quality recordings, the proposed system, equipped with the UniDepth model, localizes 87\% of all theoretically visible signs. Notably, we detect more signs than are currently recorded in the city's database. The existing database omits nearly all temporarily mounted signs and exhibits a coordinate deviation of over two meters. UniDepth performed best for signs located within 20m of the camera, beyond which its performance is on par with DepthPro. With the crowdsourced Mapillary data, locating all theoretically visible signs is more challenging, although most can still be found. Once again, UniDepth yields the best overall results, particularly at shorter distance intervals.
\begin{figure}
    \centering
    \includegraphics[width=0.9\linewidth]{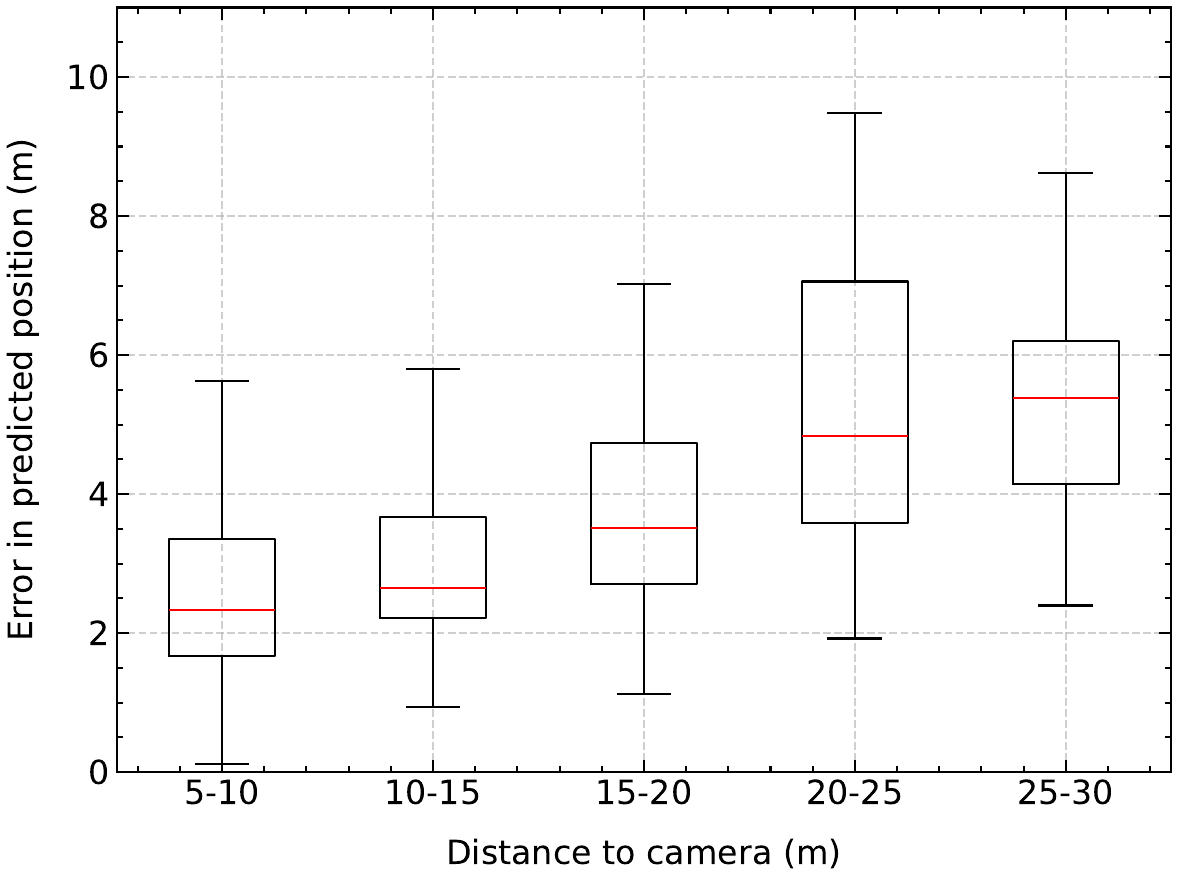}
    \caption{Relationship between true distance (camera-sign) and position error for matched predictions using UniDepth.}
    \label{fig:sign_distributions}
\end{figure}
Figure \ref{fig:sign_distributions} presents the coordinate deviation of the UniDepth model across finer-grained camera-to-sign distance intervals for all matched predictions. As illustrated, the positional error remains notably low for objects in close proximity to the camera, but the deviation increases significantly once the ground-truth distance exceeds 20 meters.

%This non-linear error growth reinforces our earlier recommendation to restrict the mapping radius to near-field objects in order to maintain high spatial reliability for the database.

\subsection{Case Study 2: Road Pavement Damage}

Road pavement damage typically appears at the bottom of the image frame and is situated closer to the camera than traffic signs. We annotated 527 instances of damage across 112 images. For the objects detected in these images, we used our Map Anything Framework to estimate their geo-coordinates and compared them with the ground-truth coordinates.

Table \ref{tab:case_studies} (c) shows that DepthAnything yields the smallest deviation, followed by UniDepth. The localization errors were generally lower than those observed for traffic signs, likely because road pavement damage is closer to the camera, with an average distance of 7.55 meters. This supports our finding that distance estimation is more accurate for nearer objects. UniDepth performs best for damage located within six meters, while DepthAnything is superior for distances exceeding six meters.

\section{Conclusion}
\label{sec:conclusion}

In this study, we systematically evaluated MapAnything, a highly scalable, low-cost pipeline that uses monocular metric depth estimation and standard camera geometry to extract 3D geo-coordinates of urban assets from a single 2D image. By validating four SOTA depth models against high-precision LiDAR data, we demonstrated that standard imagery can effectively support the budget-friendly, continuous maintenance of urban digital twins without relying on expensive, specialized hardware.

Furthermore, this evaluation highlights the pipeline's potential to detect temporal changes in the urban environment. By continuously processing imagery from varying dates, city administrations can automatically log spatial discrepancies to flag missing or damaged assets. Whether processing routine municipal street recordings or integrating with citizen reporting platforms, this approach offers an automated, accessible tool to maintain continuously updated spatial overviews of urban infrastructure.

\section{Discussion}
\label{sec:discussion}

Consultations with local municipal experts revealed that the tolerable spatial error for urban assets depends largely on object density and classification granularity. For instance, densely distributed traffic signs with high class variance are more easily matched to existing database entries than sparsely distributed items, as these factors independently reduce ambiguity. Furthermore, practical insights indicate that even manually measured coordinates in existing municipal databases can deviate by up to five meters while retaining substantial operational value. Based on these practical constraints, our evaluation demonstrates that the MapAnything pipeline is highly effective for mapping objects within a 20-meter distance, where spatial deviations consistently remain below this five-meter operational threshold.

While monocular depth estimation offers a cost-effective, high-frequency alternative to infrequent LiDAR mapping, several practical limitations remain. Real-world street-view imagery is often subject to occlusions, motion blur, and adverse environmental conditions (e.g., poor lighting, rain), which can degrade the accuracy of depth estimation. Future work must systematically assess the precise impact of these environmental factors, alongside inherent sensor inaccuracies in crowd-sourced data, such as minor heading or yaw deviations, on the final coordinate calculations. Finally, while the inference times of the evaluated models (e.g., 0.4s for UniDepth-S) demonstrate near real-time potential, scaling this pipeline for digital twin integration will require standardized spatial data models (e.g., CityGML) and robust spatial indexing to ensure seamless interoperability.
\section*{Acknowledgement}

The authors acknowledge the financial support by the Federal Ministry of Research, Technology and Space of Germany and by Sächsische Staatsministerium für Wissenschaft, Kultur und Tourismus in the program Center of Excellence for AI-research ``Center for Scalable Data Analytics and Artificial Intelligence Dresden/Leipzig'', project identification number: ScaDS.AI

We would also like to thank the German Federal Ministry for Digital and Transport for funding the DiGuRaL project through the mFund program.

\bibliography{references}

\clearpage
\setcounter{section}{0}

\onecolumn

\begin{center}
    {\titlesize\bf Supplementary Material \par}
\end{center}

\vspace{20pt}

\section{Semantic Groups for Test Images}

\begin{figure*}[htb]
    \centering
    \begin{subfigure}[b]{0.31\textwidth}
        \centering
        \includegraphics[width=\textwidth]{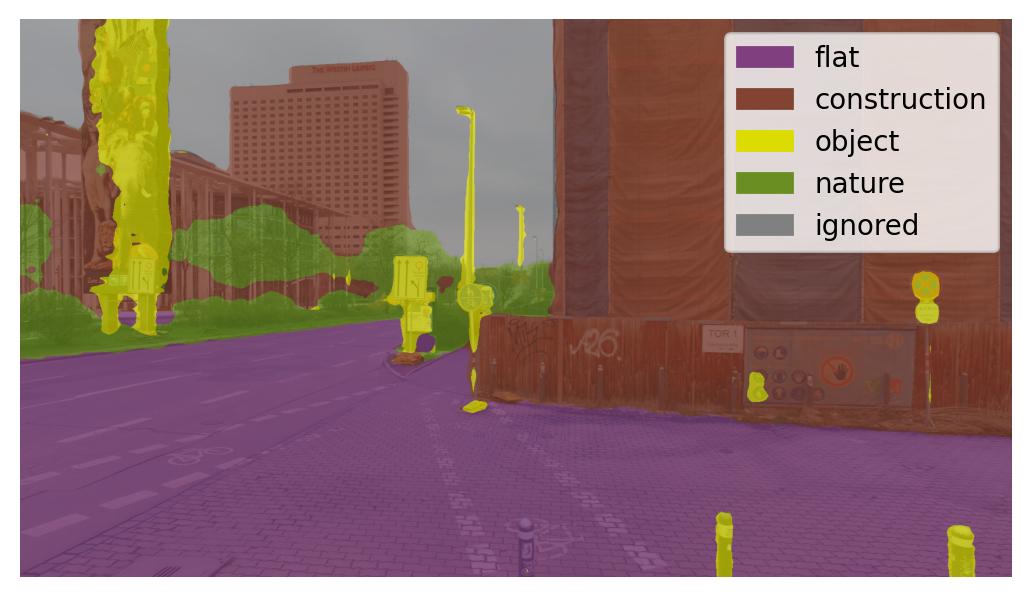}
    \end{subfigure}
    \hfill
    \begin{subfigure}[b]{0.31\textwidth}
        \centering
        \includegraphics[width=\textwidth]{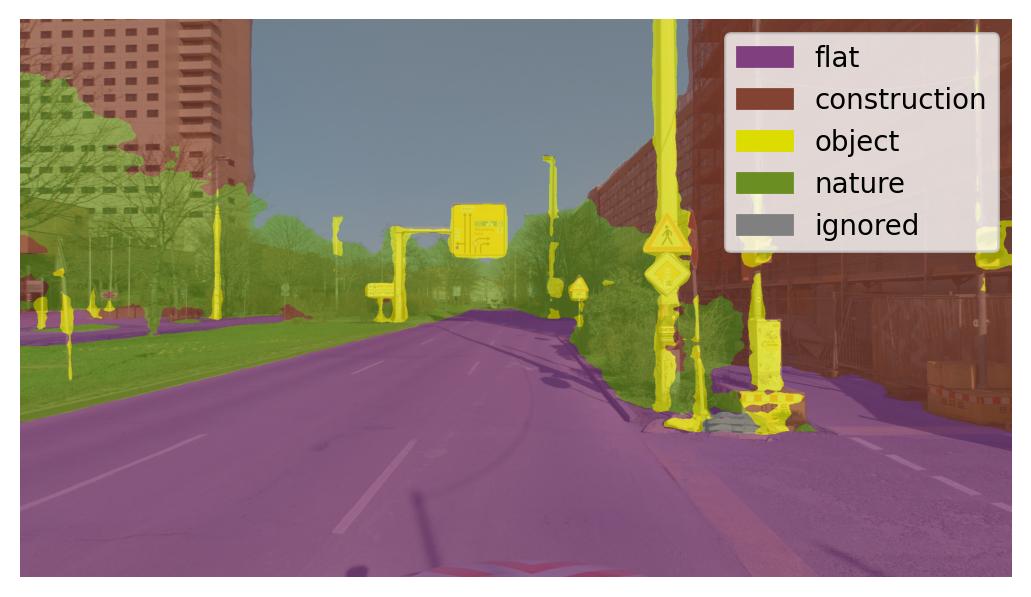}
    \end{subfigure}
    \hfill
    \begin{subfigure}[b]{0.31\textwidth}
        \centering
        \includegraphics[width=\textwidth]{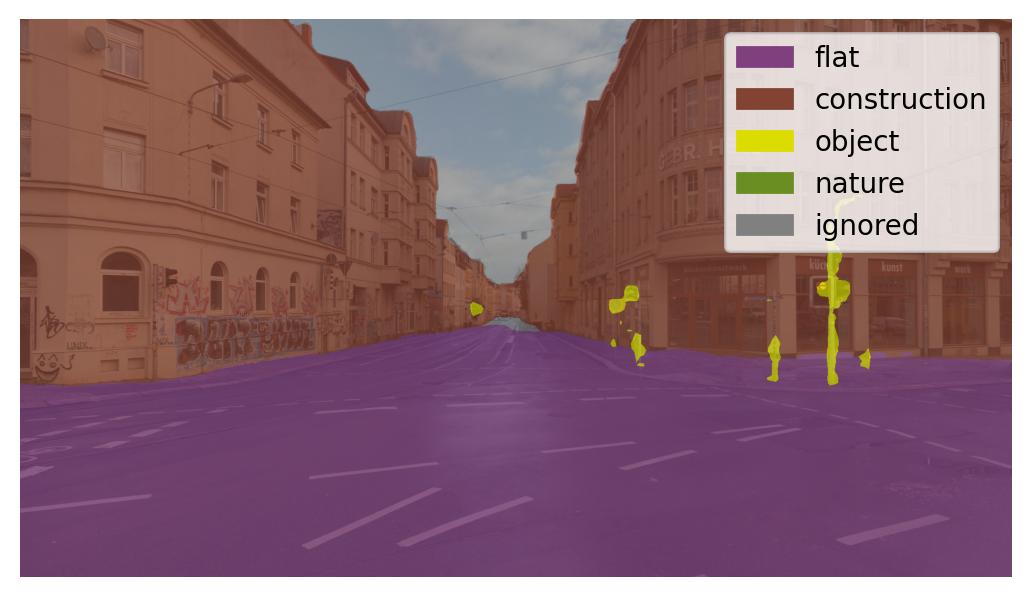}
    \end{subfigure}

    \vspace{1em}

    \begin{subfigure}[b]{0.31\textwidth}
        \centering
        \includegraphics[width=\textwidth]{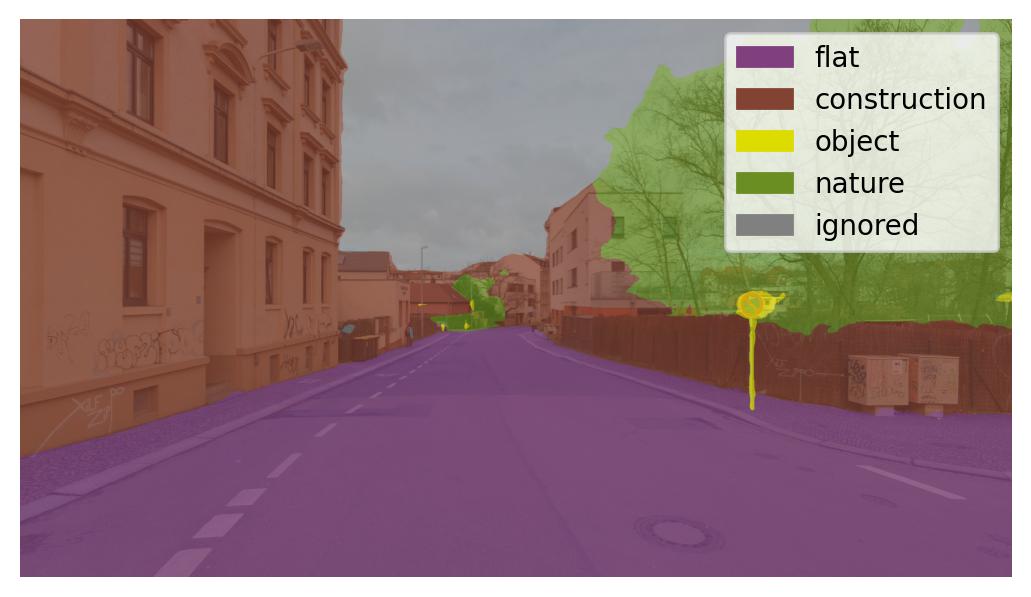}
    \end{subfigure}
    \hfill
    \begin{subfigure}[b]{0.31\textwidth}
        \centering
        \includegraphics[width=\textwidth]{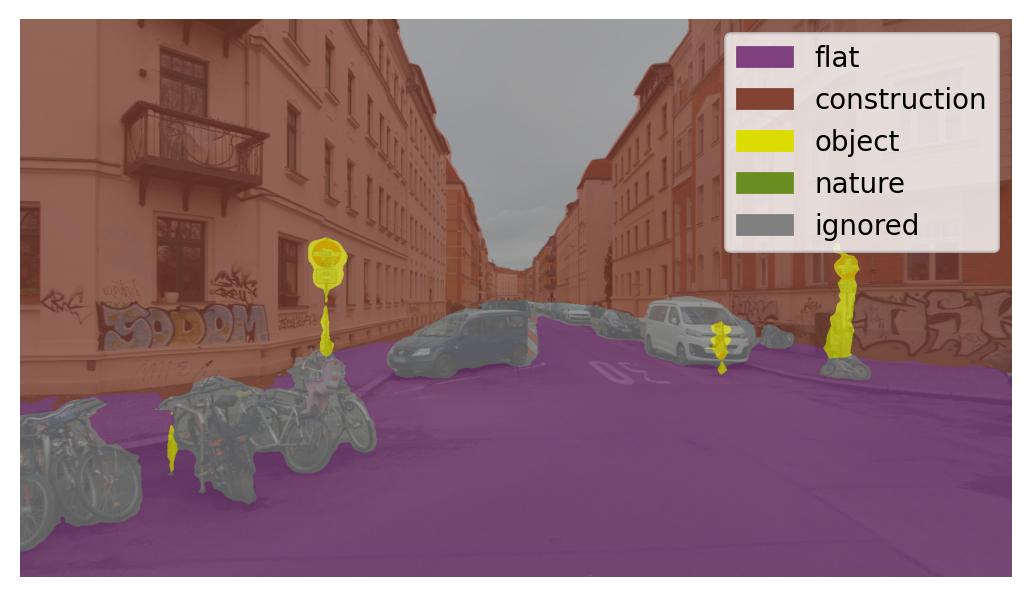}
    \end{subfigure}
    \hfill
    \begin{subfigure}[b]{0.31\textwidth}
        \centering
        \includegraphics[width=\textwidth]{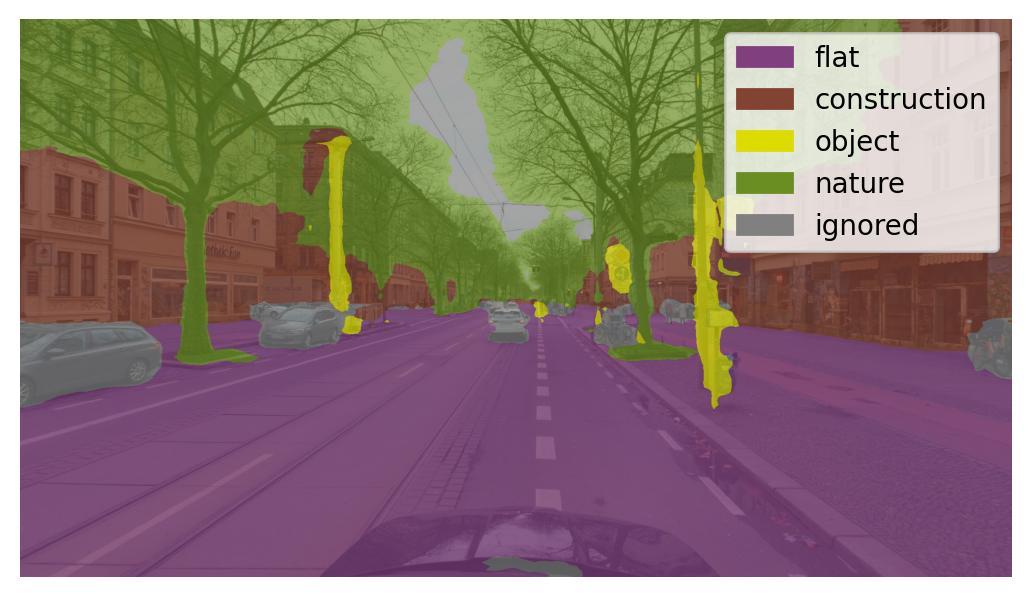}
    \end{subfigure}

    \vspace{1em}

    \begin{subfigure}[b]{0.31\textwidth}
        \centering
        \includegraphics[width=\textwidth]{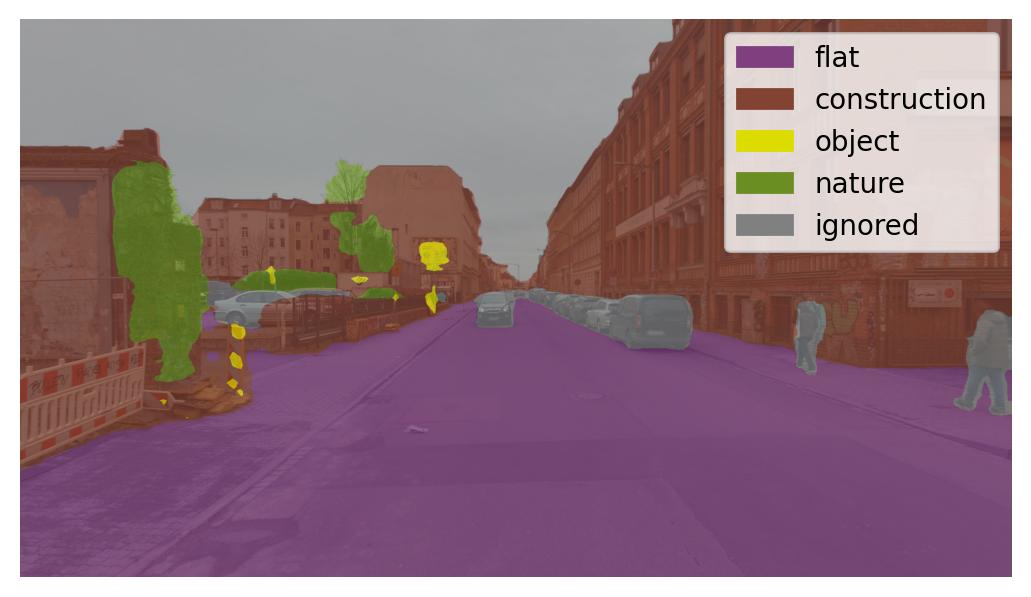}
    \end{subfigure}
    \hfill
    \begin{subfigure}[b]{0.31\textwidth}
        \centering
        \includegraphics[width=\textwidth]{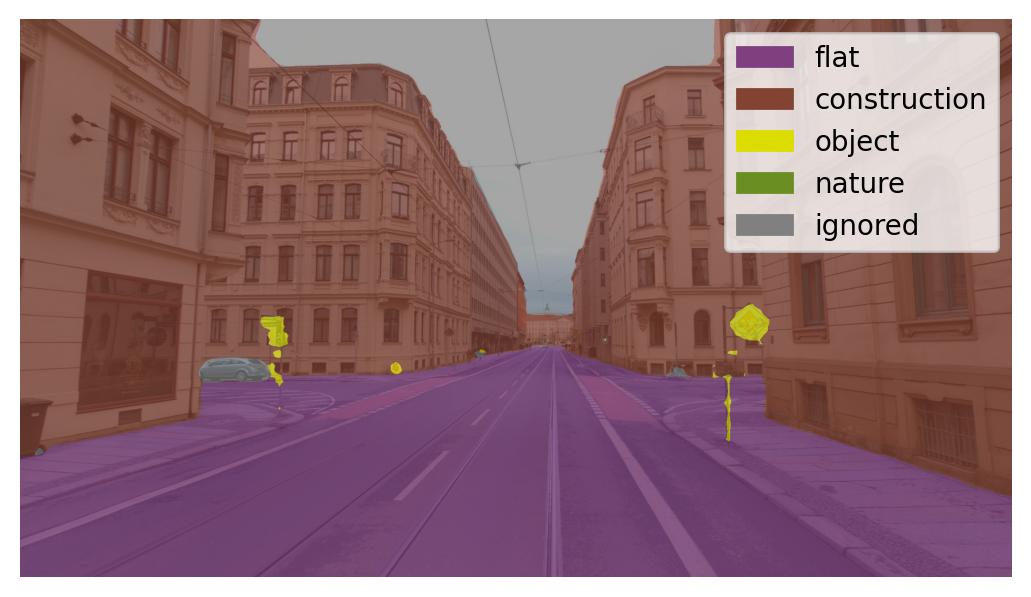}
    \end{subfigure}
    \hfill
    \begin{subfigure}[b]{0.31\textwidth}
        \centering
        \includegraphics[width=\textwidth]{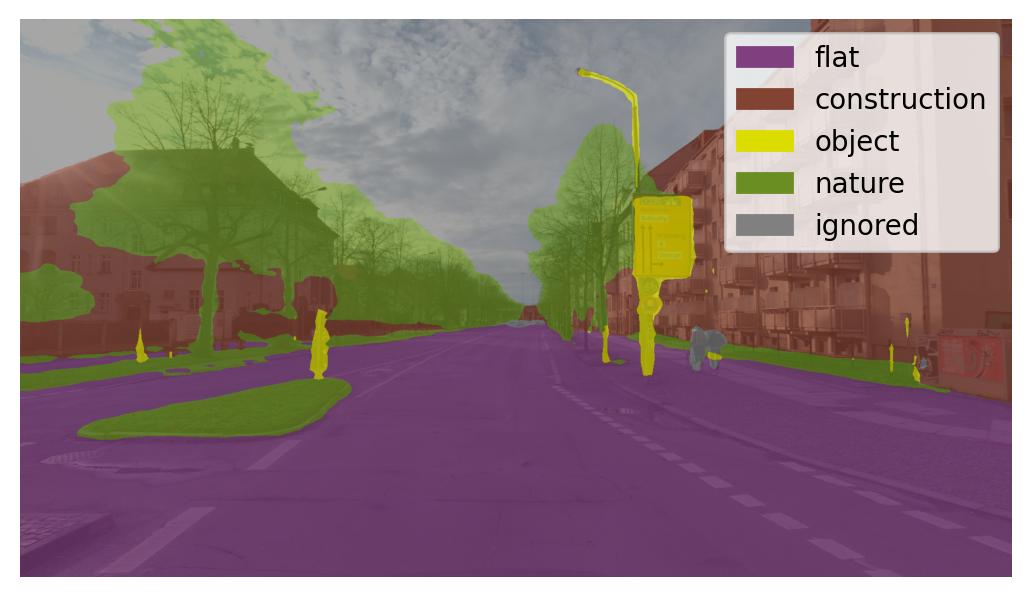}
    \end{subfigure}

    \vspace{1em}

    \begin{subfigure}[b]{0.31\textwidth}
        \centering
        \includegraphics[width=\textwidth]{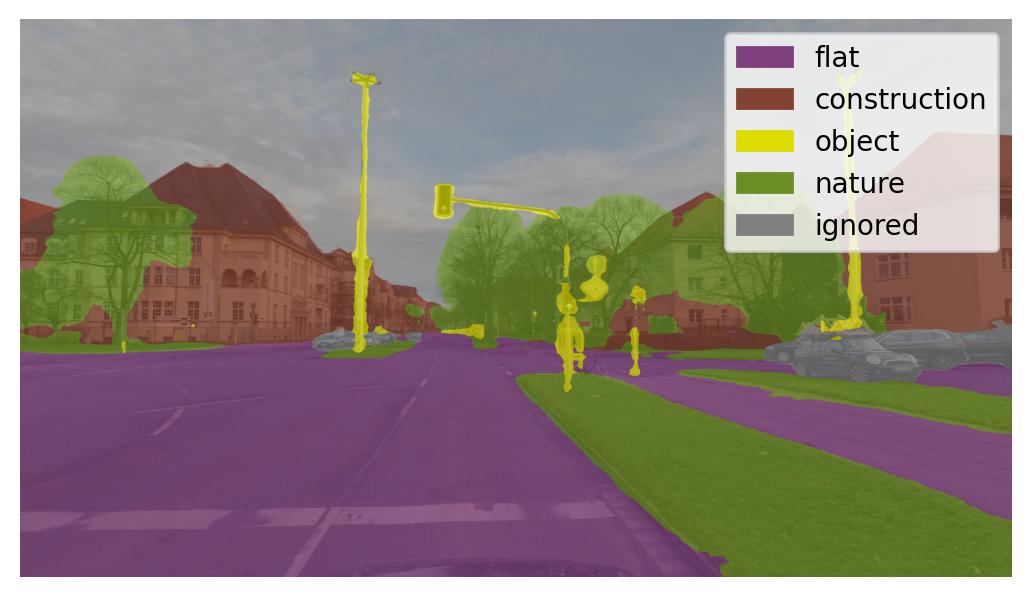}
    \end{subfigure}
    \hfill
    \begin{subfigure}[b]{0.31\textwidth}
        \centering
        \includegraphics[width=\textwidth]{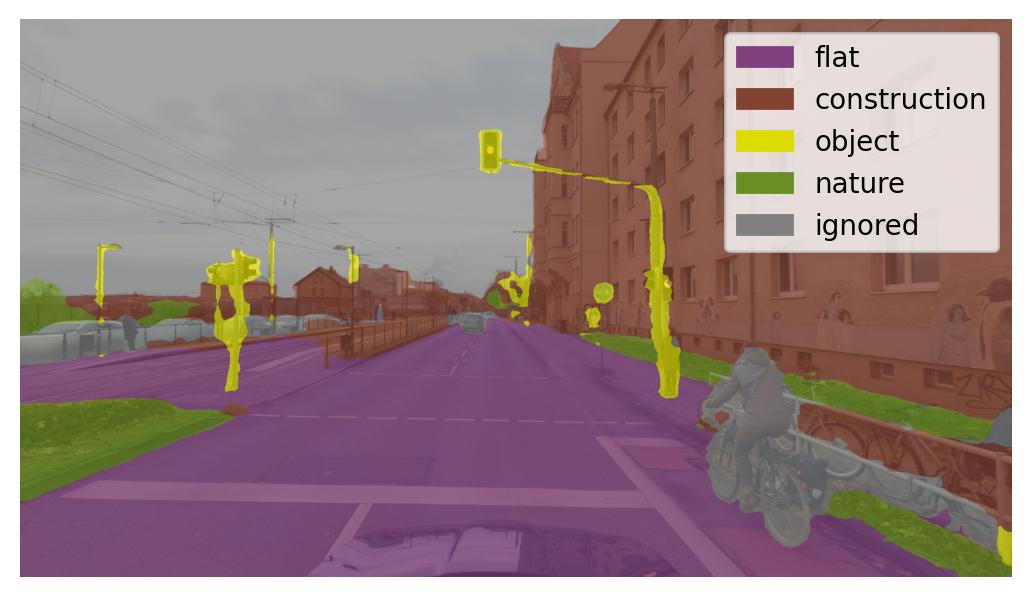}
    \end{subfigure}
    \hfill
    \begin{subfigure}[b]{0.31\textwidth}
        \centering
        \includegraphics[width=\textwidth]{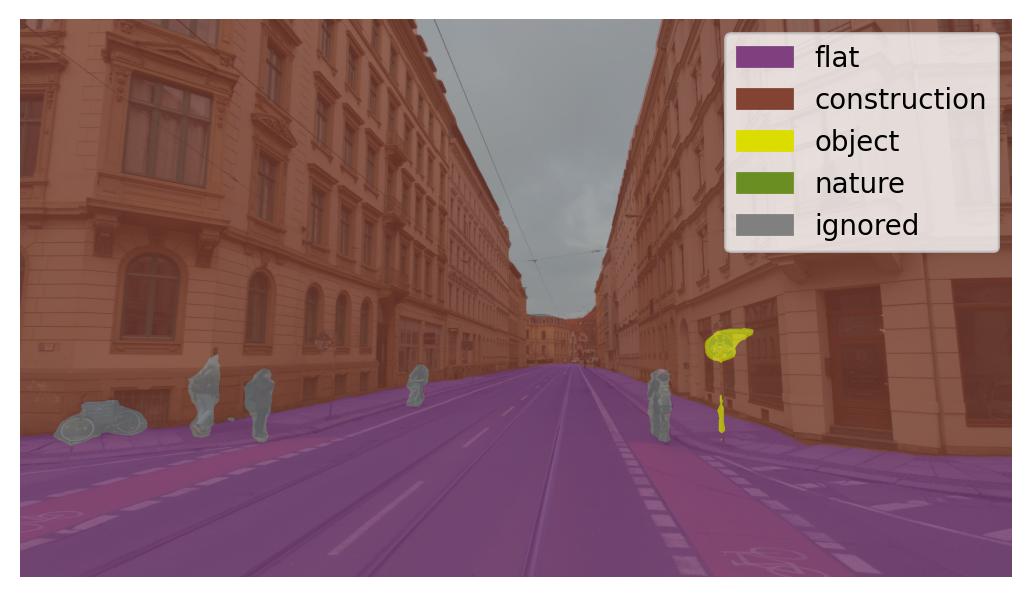}
    \end{subfigure}

    \caption{Semantic Groups for twelve of the sample images. Original images from  Cyclomedia.}
    
\end{figure*}

\clearpage
\section{All ARE plots for one Example Image}

\begin{figure*}[htb]
    \centering

    % --- Row 1: DepthAnything ---
    \begin{subfigure}[b]{0.32\textwidth}
        \centering
        \includegraphics[width=\textwidth]{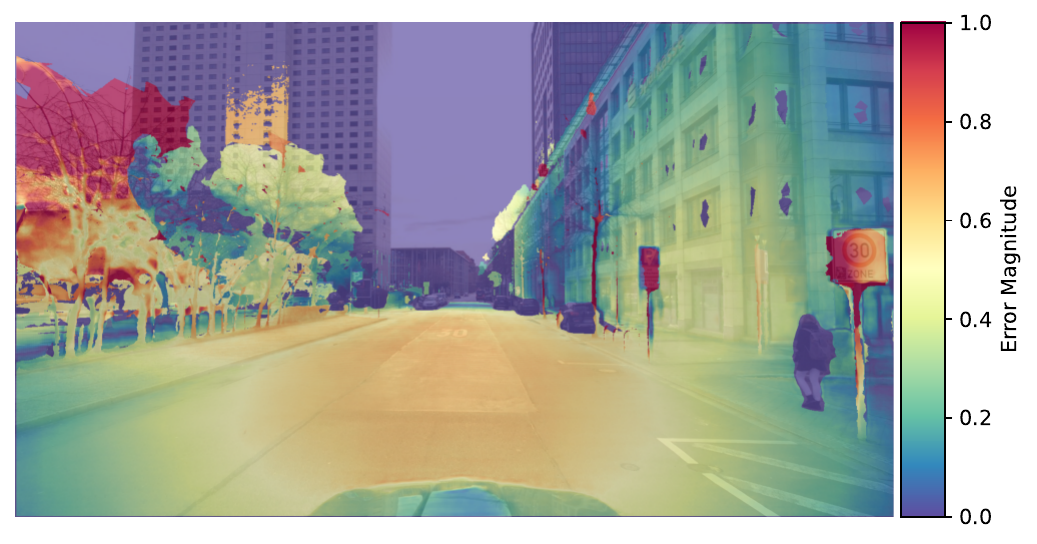}
        \caption{DepthAnything-S}
    \end{subfigure}
    \hfill
    \begin{subfigure}[b]{0.32\textwidth}
        \centering
        \includegraphics[width=\textwidth]{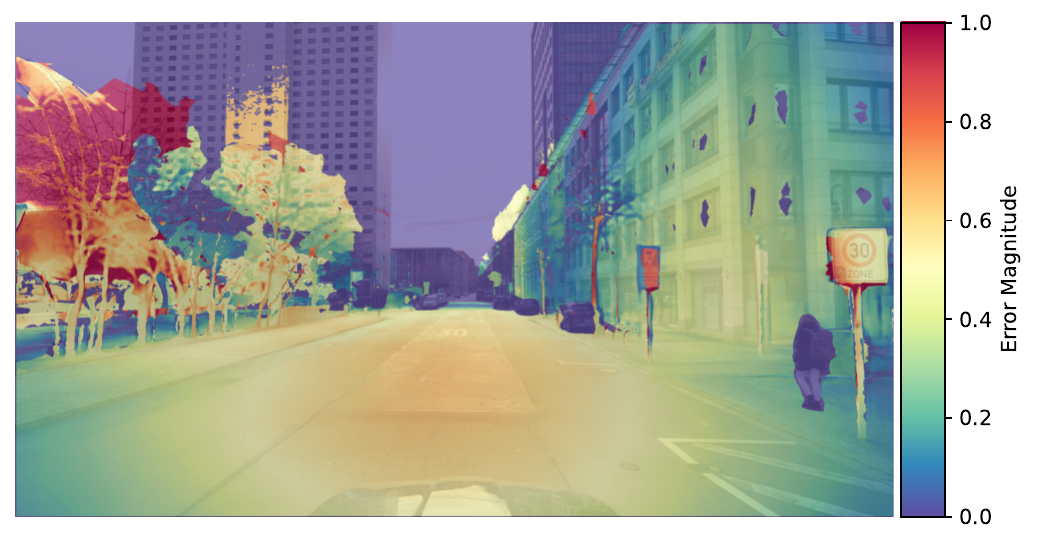}
        \caption{DepthAnything-B}
    \end{subfigure}
    \hfill
    \begin{subfigure}[b]{0.32\textwidth}
        \centering
        \includegraphics[width=\textwidth]{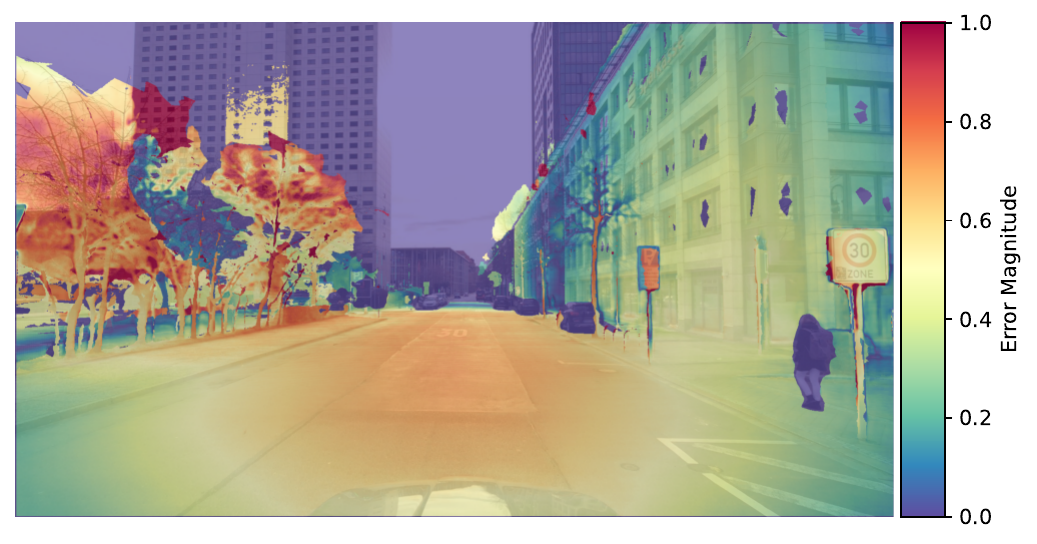}
        \caption{DepthAnything-L}
    \end{subfigure}

    \vspace{1em}

    % --- Row 2: UniDepth ---
    \begin{subfigure}[b]{0.32\textwidth}
        \centering
        \includegraphics[width=\textwidth]{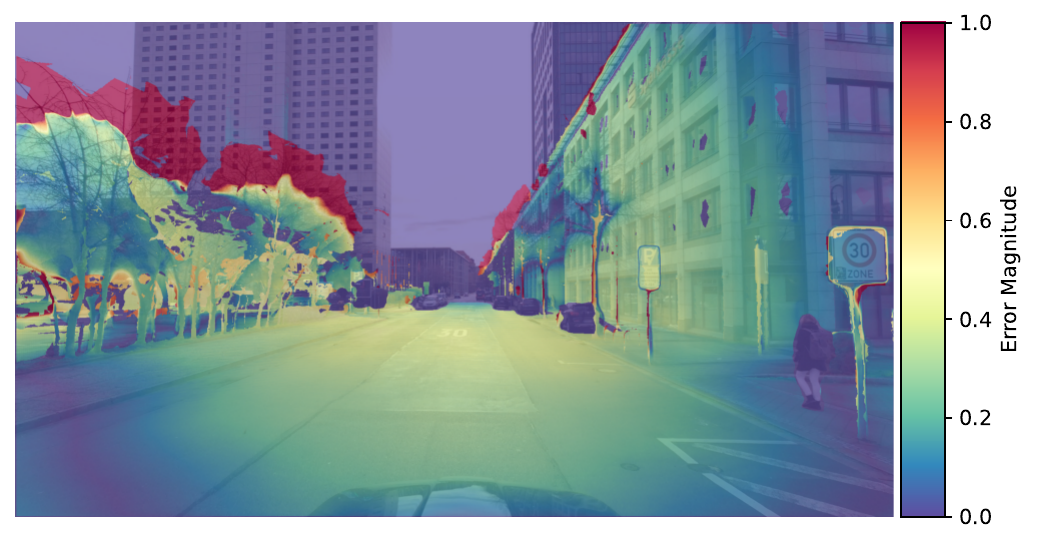}
        \caption{UniDepth-S}
    \end{subfigure}
    \hfill
    \begin{subfigure}[b]{0.32\textwidth}
        \centering
        \includegraphics[width=\textwidth]{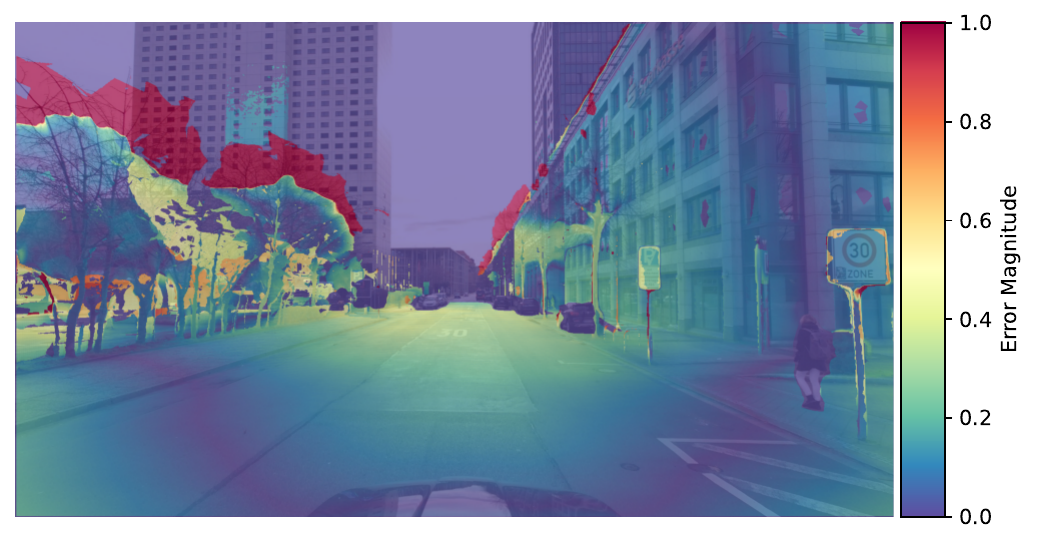}
        \caption{UniDepth-B}
    \end{subfigure}
    \hfill
    \begin{subfigure}[b]{0.32\textwidth}
        \centering
        \includegraphics[width=\textwidth]{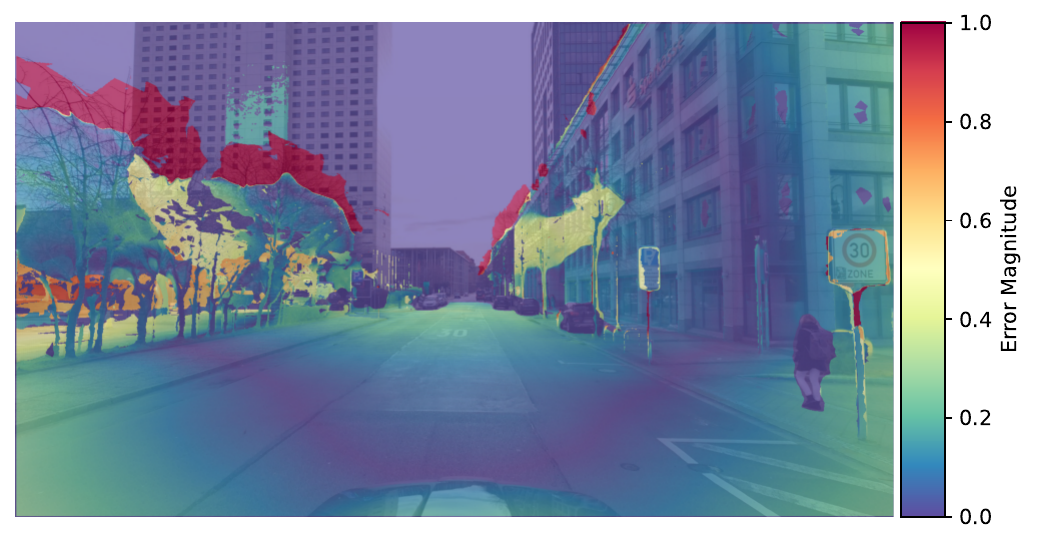}
        \caption{UniDepth-L}
    \end{subfigure}

    \vspace{1em}

    % --- Row 3: DepthPro ---
    \begin{subfigure}[b]{0.32\textwidth}
        \centering
        \includegraphics[width=\textwidth]{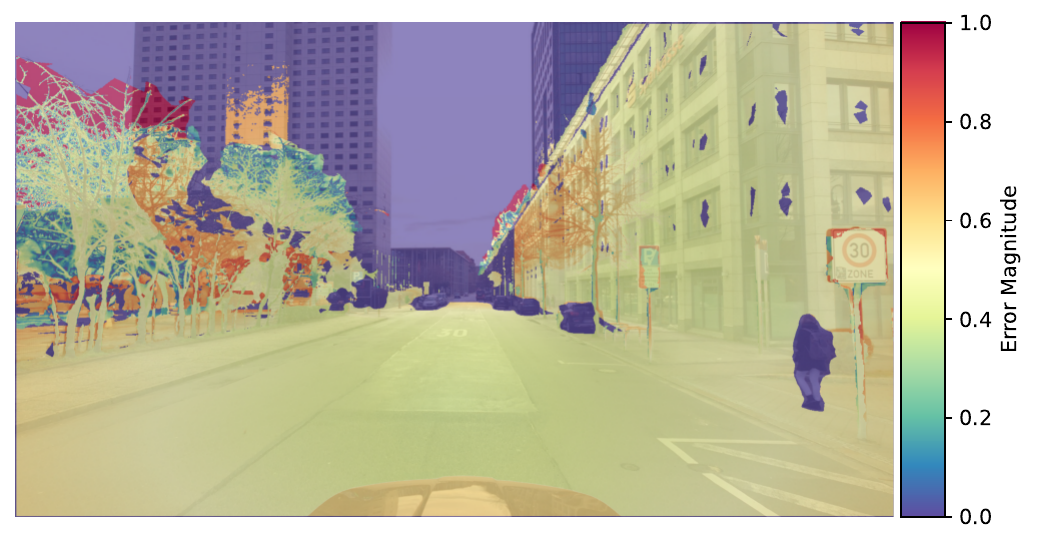}
        \caption{DepthPro (with cam)}
    \end{subfigure}
    \hfill
    \begin{subfigure}[b]{0.32\textwidth}
        \centering
        \includegraphics[width=\textwidth]{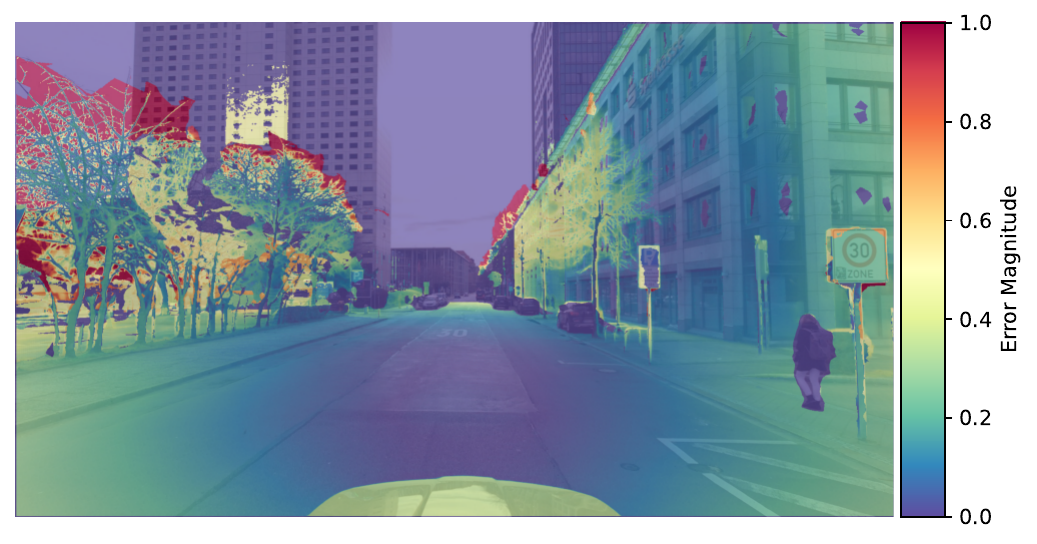}
        \caption{DepthPro (no cam)}
    \end{subfigure}
    \hfill
    \begin{subfigure}[b]{0.32\textwidth}
        \centering
        \rule{\textwidth}{0pt} 
    \end{subfigure}

    \vspace{1em}

    % --- Row 4: Metric3D ---
    \begin{subfigure}[b]{0.32\textwidth}
        \centering
        \includegraphics[width=\textwidth]{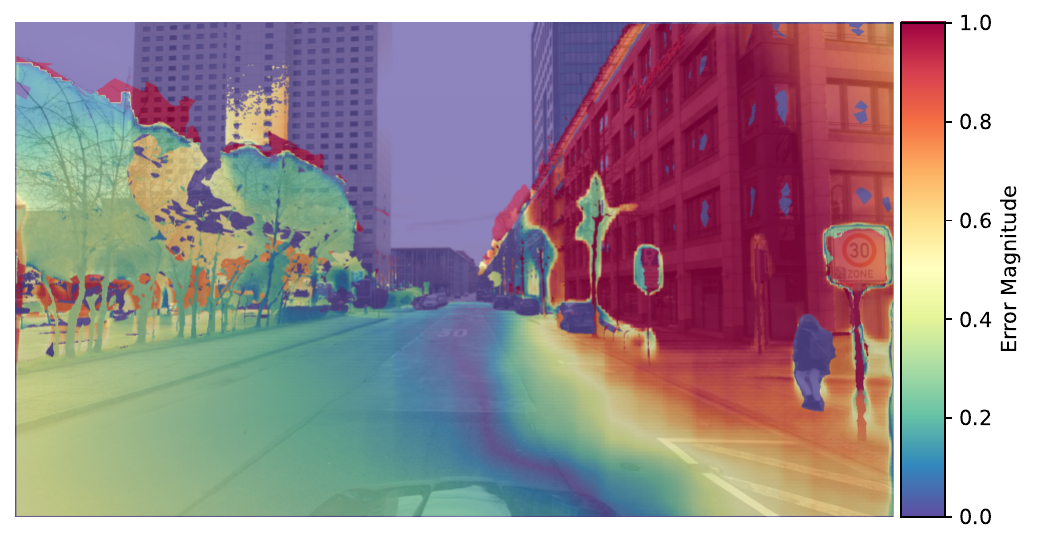}
        \caption{Metric3D-ConvNext}
    \end{subfigure}
    \hfill
    \begin{subfigure}[b]{0.32\textwidth}
        \centering
        \includegraphics[width=\textwidth]{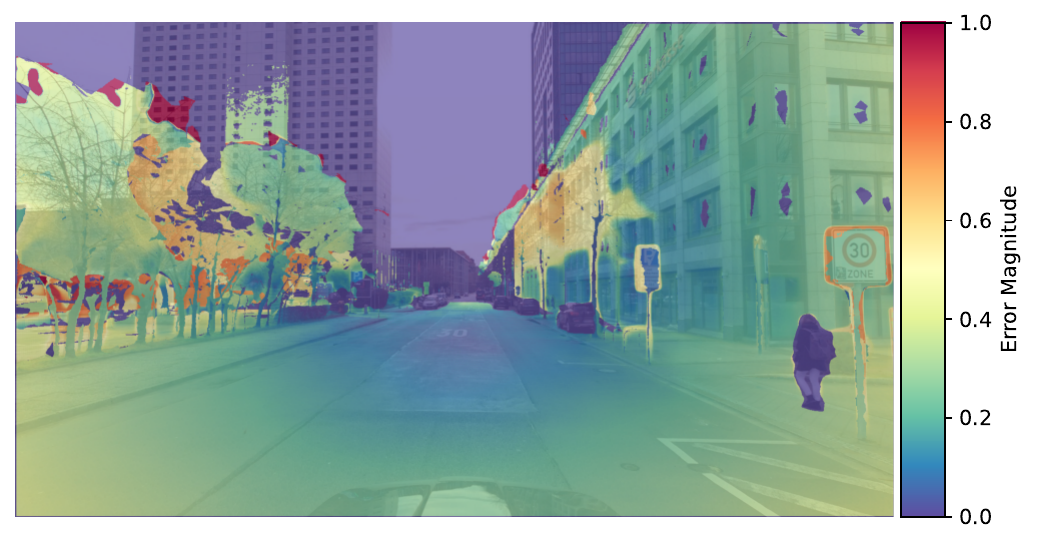}
        \caption{Metric3D-ViT}
    \end{subfigure}
    \hfill
    \begin{subfigure}[b]{0.32\textwidth}
        \centering
        \rule{\textwidth}{0pt} 
    \end{subfigure}

    \caption{Absolute Relative Error (ARE) for different depth estimation models on the same example image.}
    \label{fig:are_plots}
\end{figure*}

\clearpage
\section{All MAE plots for one Example Image}

\begin{figure*}[h!]
    \centering

    % --- Row 1: DepthAnything ---
    \begin{subfigure}[b]{0.32\textwidth}
        \centering
        \includegraphics[width=\textwidth]{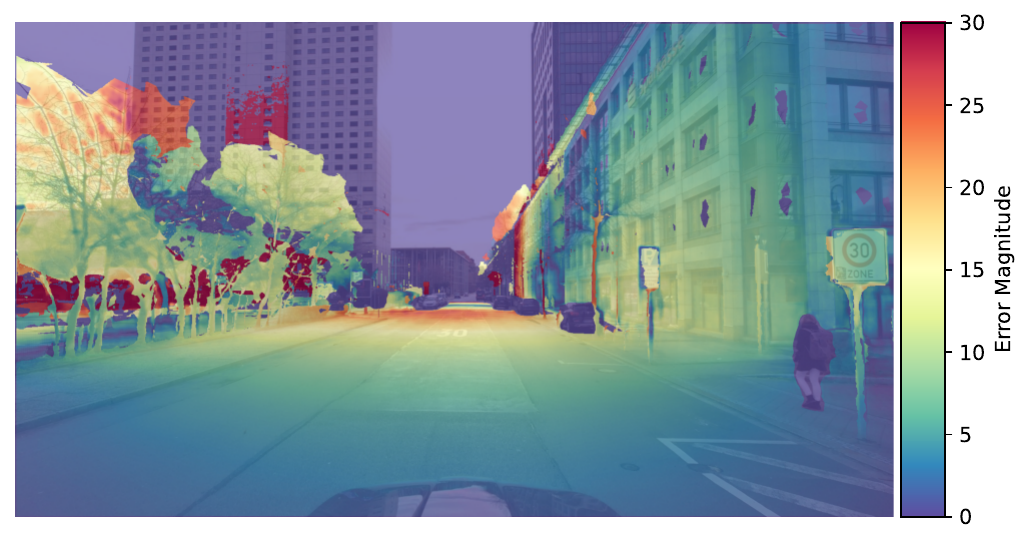}
        \caption{DepthAnything-S}
    \end{subfigure}
    \hfill
    \begin{subfigure}[b]{0.32\textwidth}
        \centering
        \includegraphics[width=\textwidth]{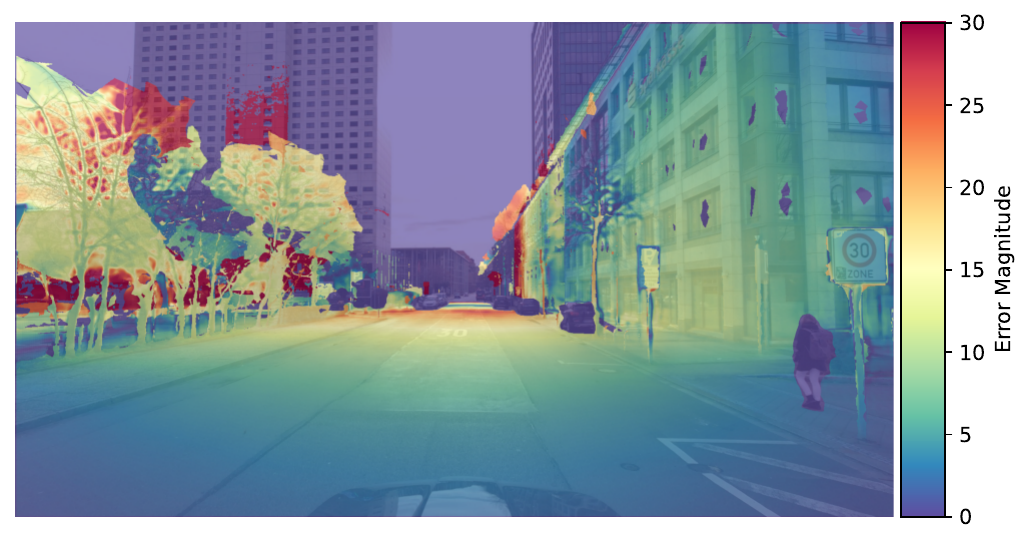}
        \caption{DepthAnything-B}
    \end{subfigure}
    \hfill
    \begin{subfigure}[b]{0.32\textwidth}
        \centering
        \includegraphics[width=\textwidth]{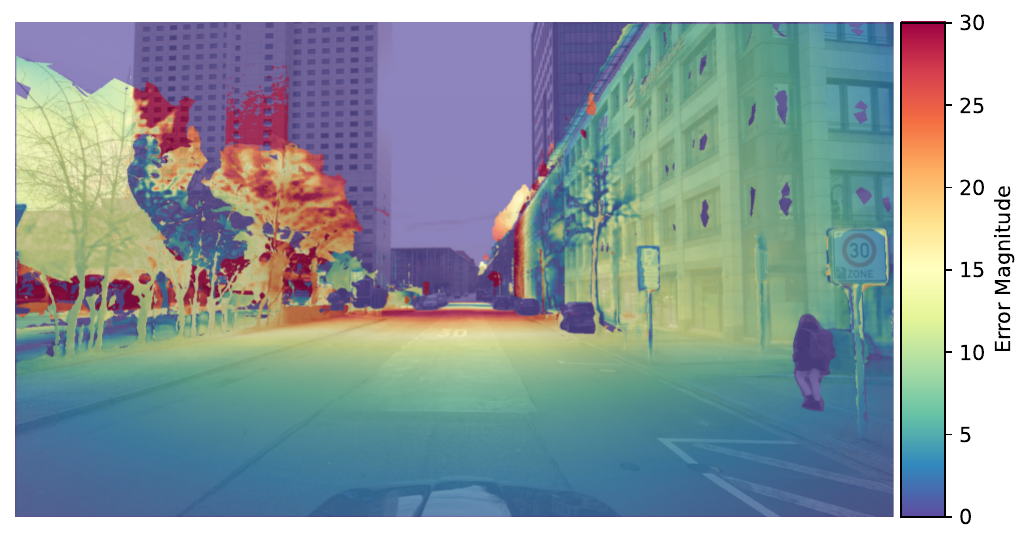}
        \caption{DepthAnything-L}
    \end{subfigure}

    \vspace{1em}

    % --- Row 2: UniDepth ---
    \begin{subfigure}[b]{0.32\textwidth}
        \centering
        \includegraphics[width=\textwidth]{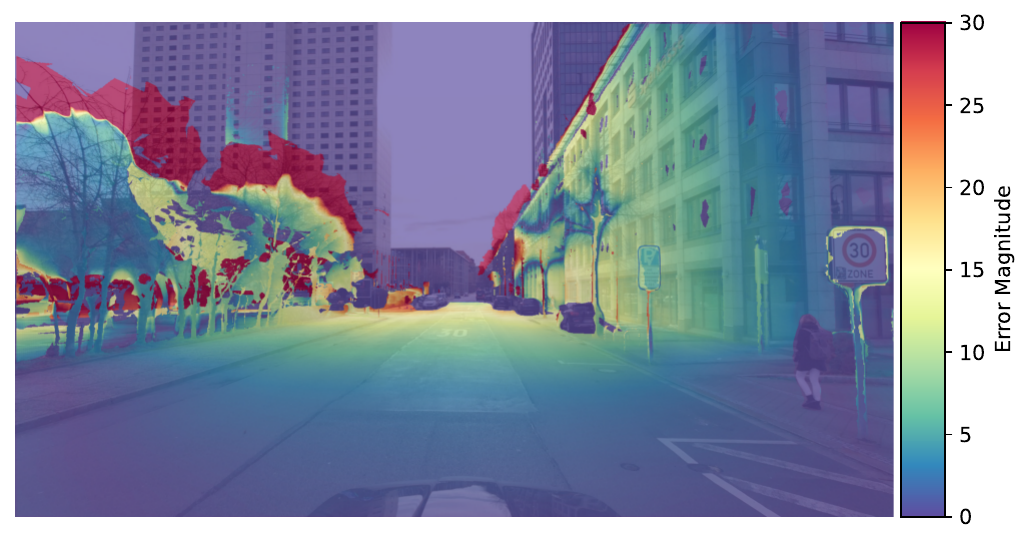}
        \caption{UniDepth-S}
    \end{subfigure}
    \hfill
    \begin{subfigure}[b]{0.32\textwidth}
        \centering
        \includegraphics[width=\textwidth]{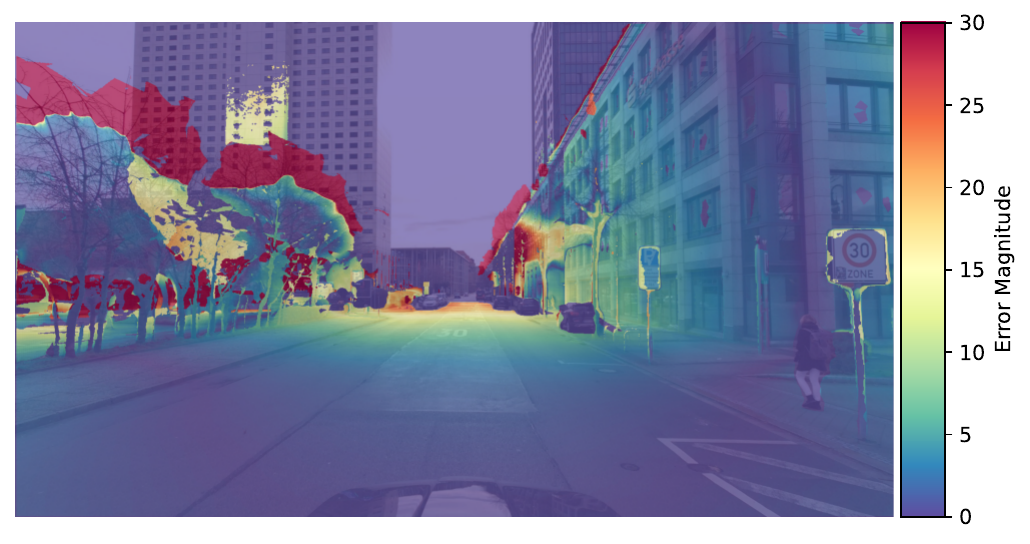}
        \caption{UniDepth-B}
    \end{subfigure}
    \hfill
    \begin{subfigure}[b]{0.32\textwidth}
        \centering
        \includegraphics[width=\textwidth]{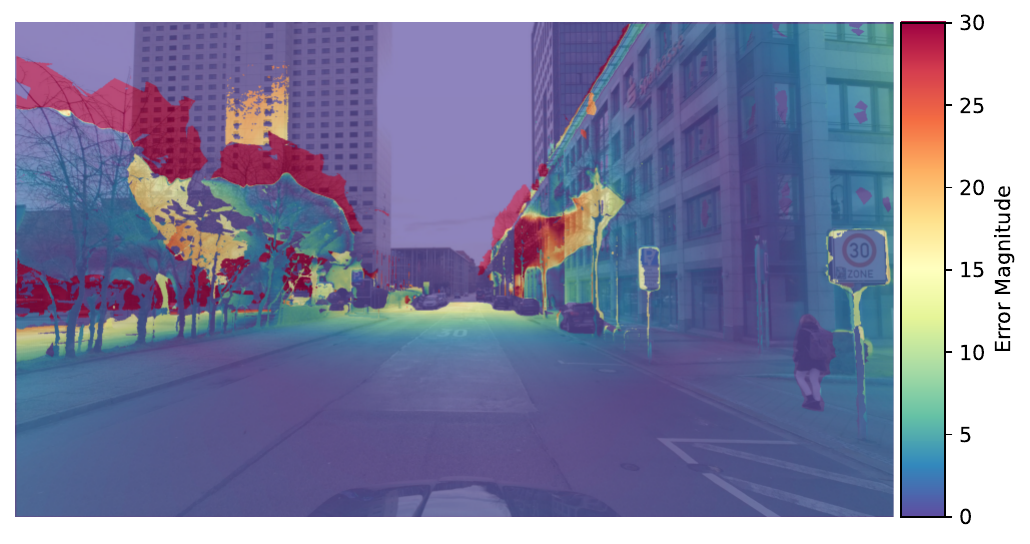}
        \caption{UniDepth-L}

    \end{subfigure}

    \vspace{1em}
    
    % --- Row 3: DepthPro ---
    \begin{subfigure}[b]{0.32\textwidth}
        \centering
        \includegraphics[width=\textwidth]{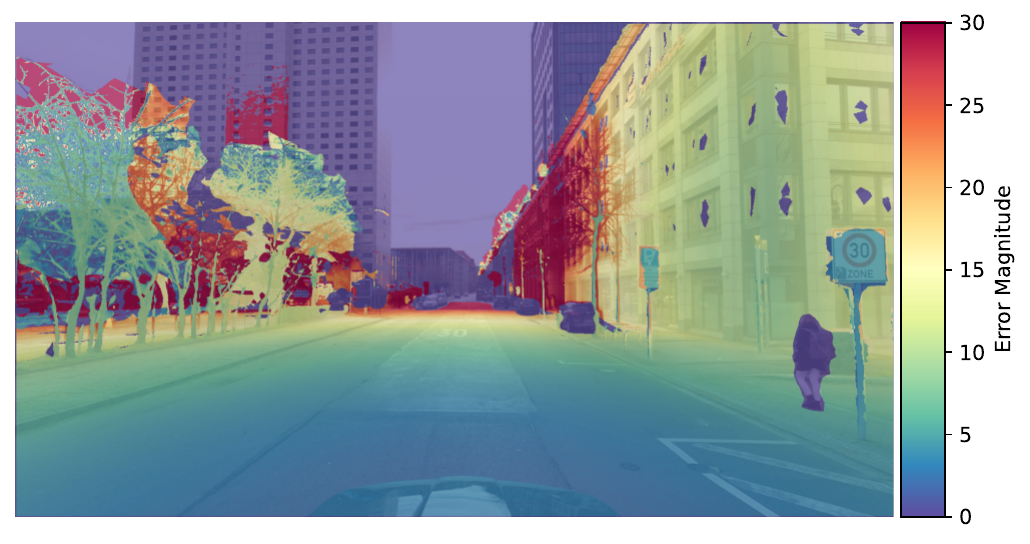}
        \caption{DepthPro (with cam)}
    \end{subfigure}
    \hfill
    \begin{subfigure}[b]{0.32\textwidth}
        \centering
        \includegraphics[width=\textwidth]{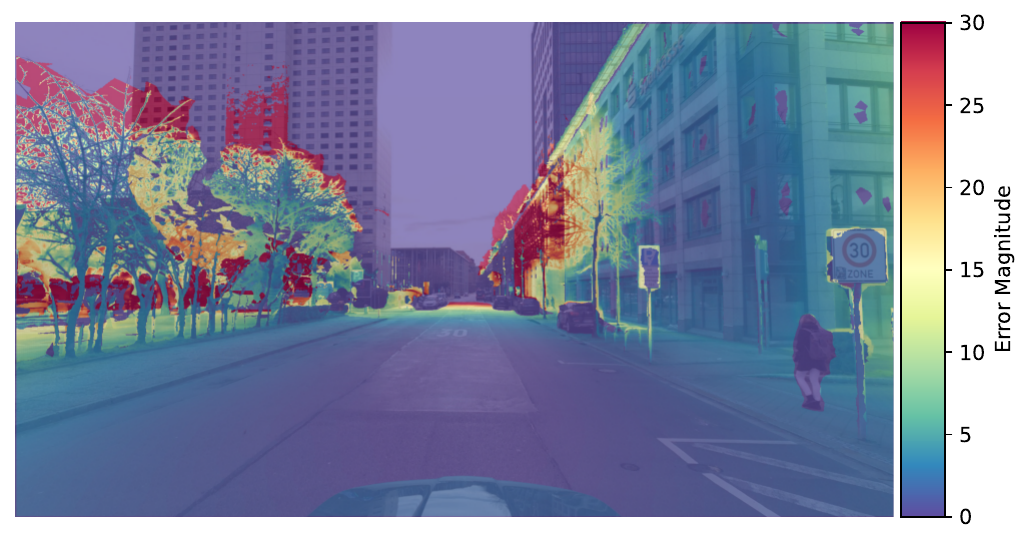}
        \caption{DepthPro (no cam)}
    \end{subfigure}
    \hfill
    \begin{subfigure}[b]{0.32\textwidth}
        \centering
        \rule{\textwidth}{0pt} 
    \end{subfigure}
    
    \vspace{1em}

    % --- Row 4: Metric3D ---
    \begin{subfigure}[b]{0.32\textwidth}
        \centering
        \includegraphics[width=\textwidth]{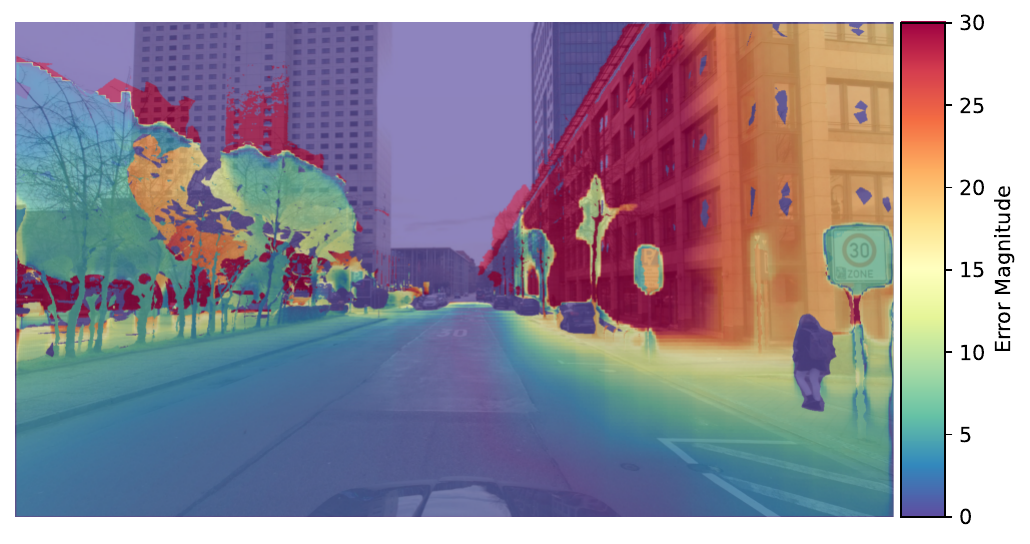}
        \caption{Metric3D-ConvNext}
    \end{subfigure}
    \hfill
    \begin{subfigure}[b]{0.32\textwidth}
        \centering
        \includegraphics[width=\textwidth]{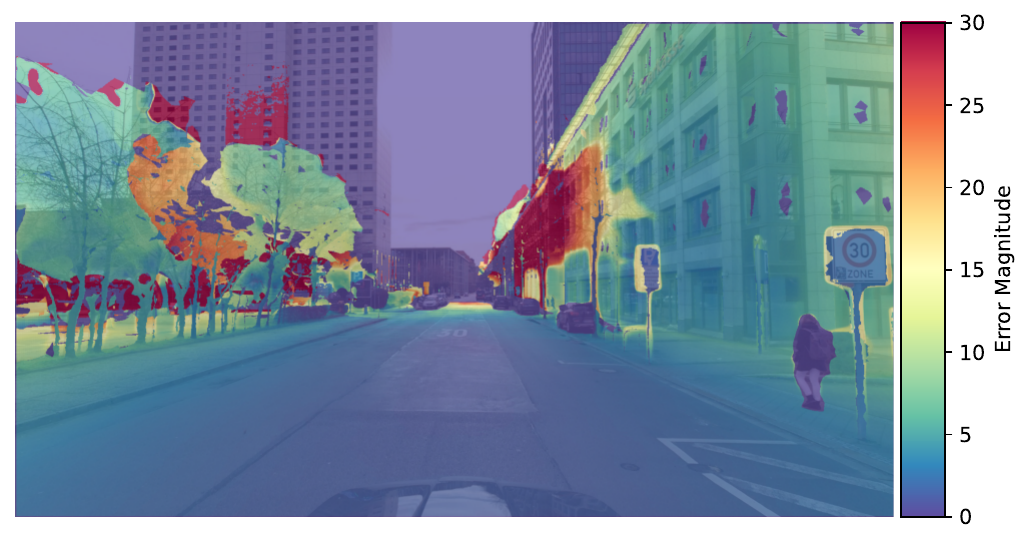}
        \caption{Metric3D-ViT}
    \end{subfigure}
    \hfill
    \begin{subfigure}[b]{0.32\textwidth}
        \centering
        \rule{\textwidth}{0pt} 
    \end{subfigure}

    \caption{Mean Absolute Error (MAE) for different depth estimation models on the same example image.}
    \label{fig:MAE_plots}
\end{figure*}

\clearpage
\section{Traffic Signs: Image Coverage in Annotated Area}

\begin{figure*}[!ht]
    \centering
    
    \begin{subfigure}[b]{0.4\textwidth}
        \centering
        \includegraphics[width=\textwidth]{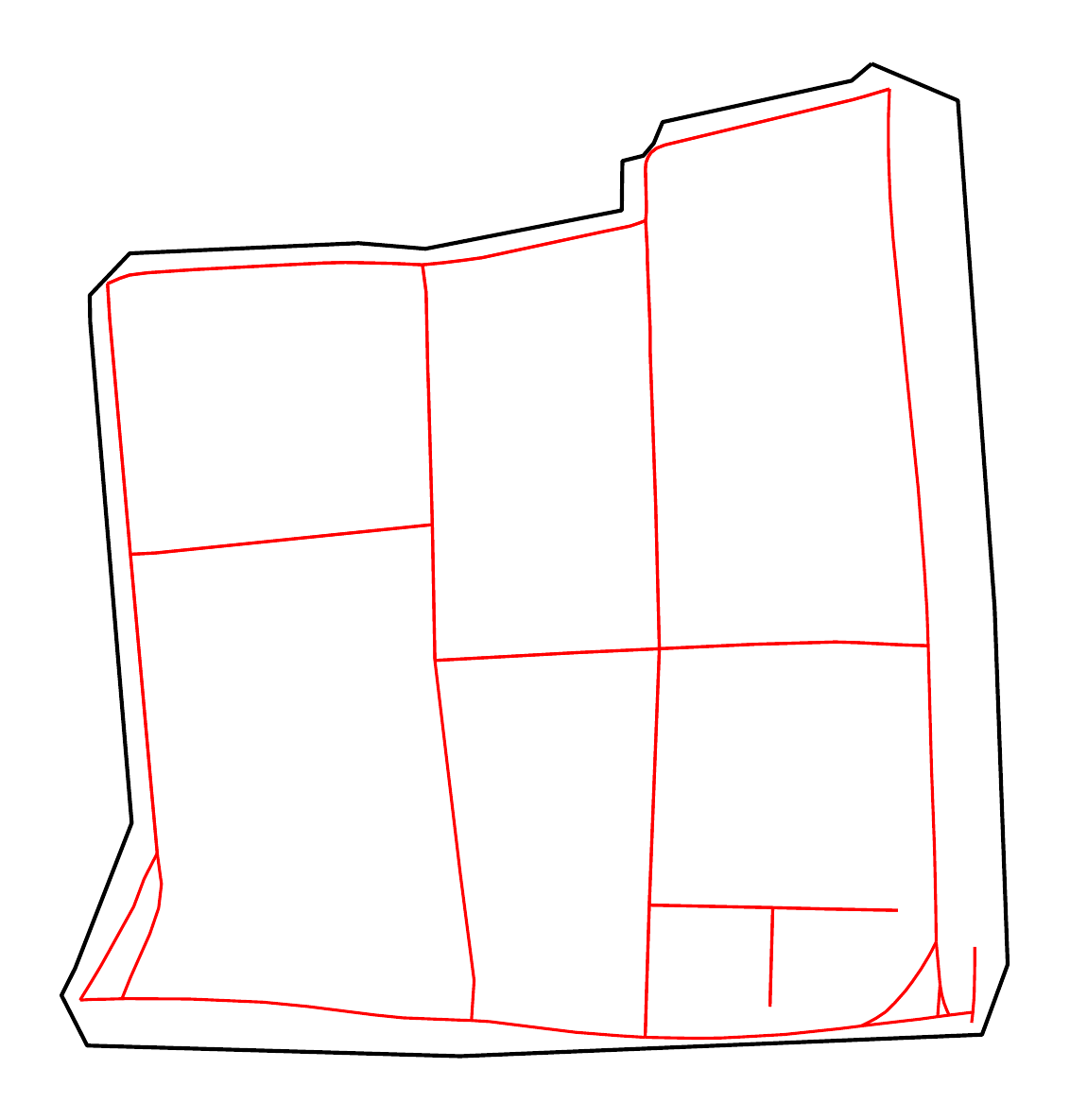}
        \caption{Road Network}
    \end{subfigure}
    
    \vspace{1em}
    
    \begin{subfigure}[b]{0.45\textwidth}
        \centering
        \includegraphics[width=\textwidth]{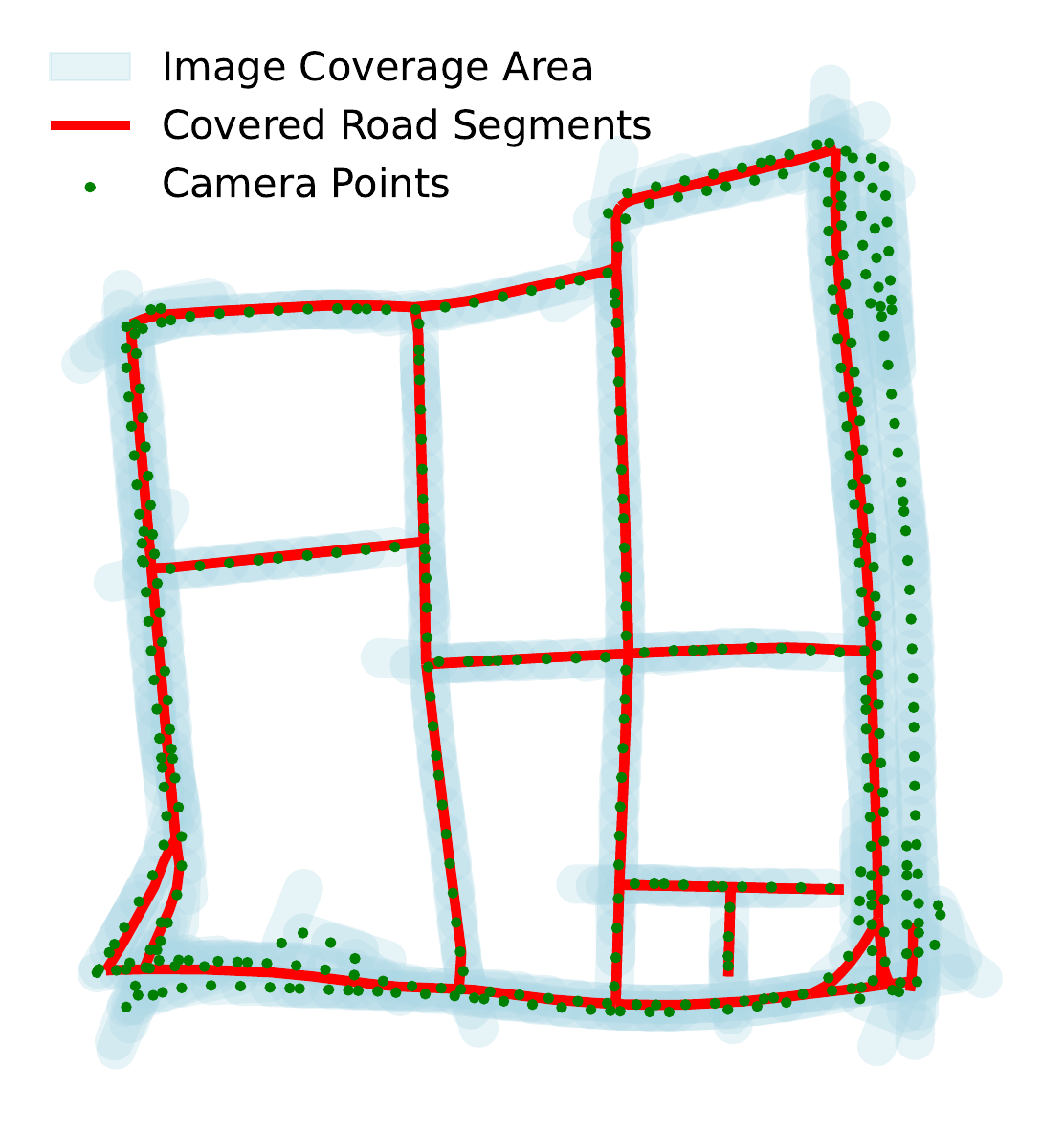}
        \caption{Road covered by Cyclomedia Images}
    \end{subfigure}
    \hfill
    \begin{subfigure}[b]{0.45\textwidth}
        \centering
        \includegraphics[width=\textwidth]{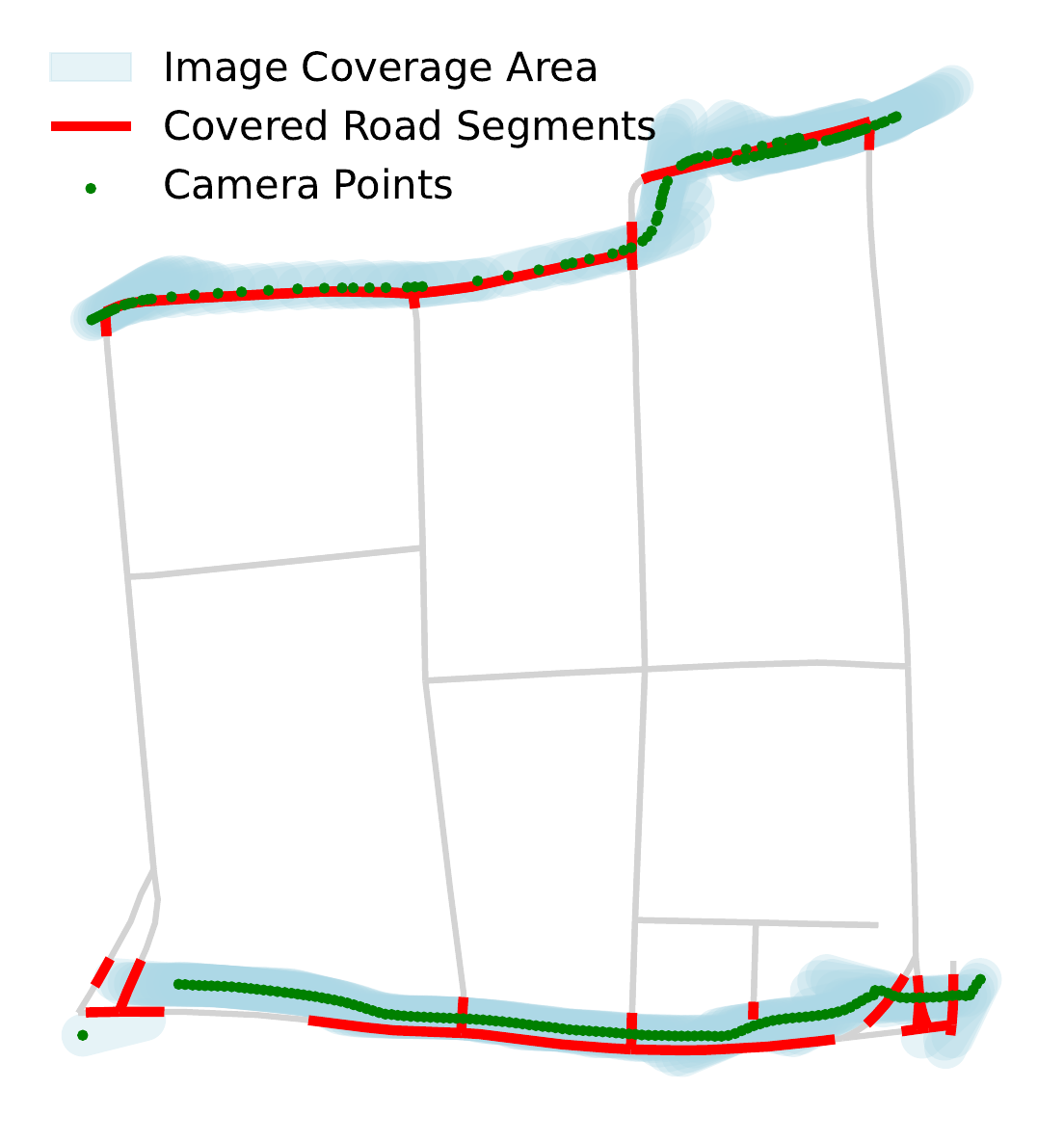}
        \caption{Road covered by Mapillary Images}
    \end{subfigure}
    
    \caption{The road network according to OpenStreetMap in the area where signs were annotated (a) and the parts of it that are covered with images from Cyclomedia (b) and Mapillary (c), including the positions of the camera. The image coverage area is assumed to be 30 meters in recording direction with a 10-meter buffer on both sides. Best viewed in colors.}
    \label{fig:image_coverage1}
\end{figure*}

\begin{figure*}[ht]
    \centering
    
    \begin{subfigure}[b]{0.45\textwidth}
        \centering
        \includegraphics[width=\textwidth]{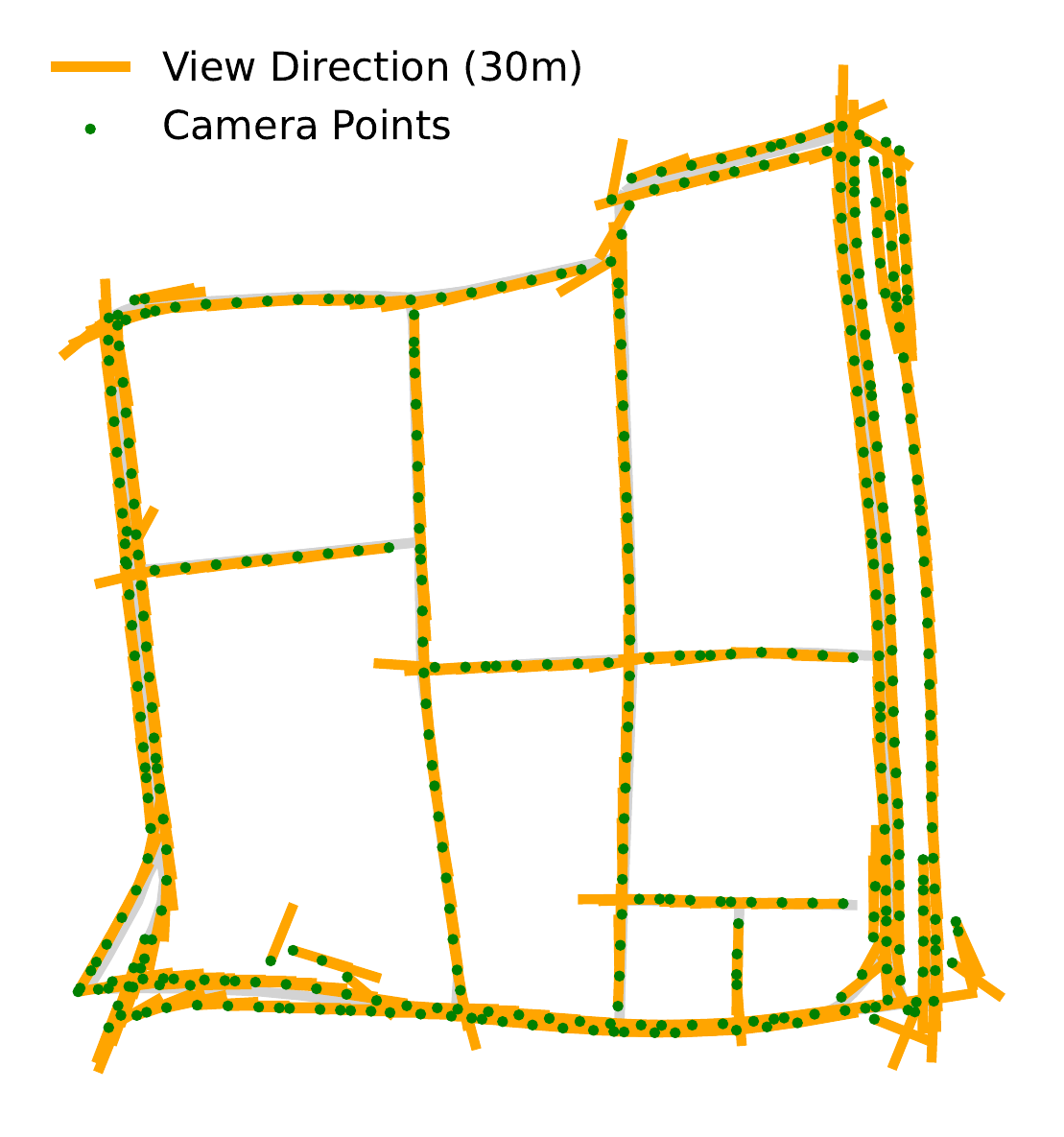}
        \caption{Camera positions and directions by Cyclomedia Images}
    \end{subfigure}
    \hfill
    \begin{subfigure}[b]{0.45\textwidth}
        \centering
        \includegraphics[width=\textwidth]{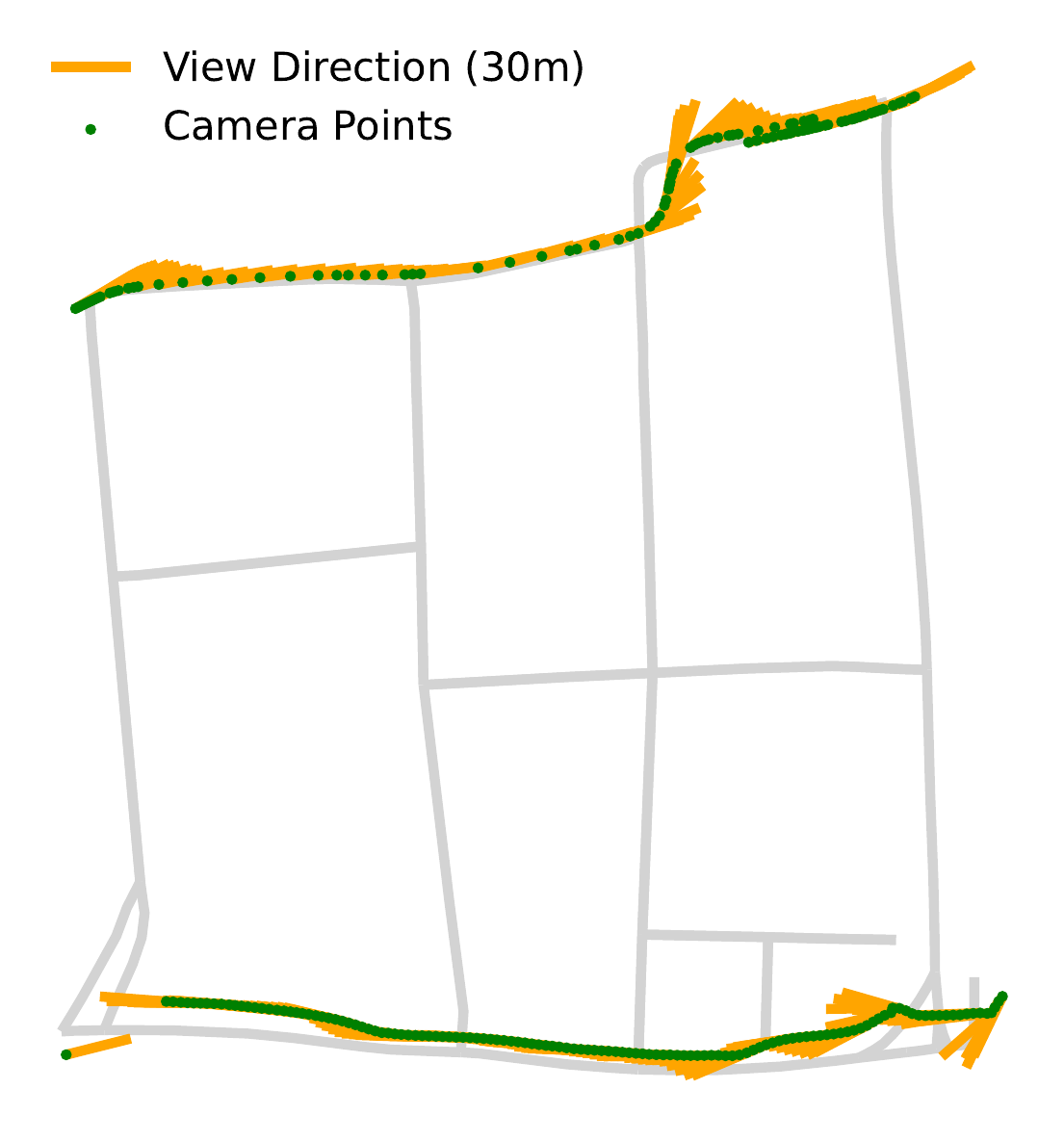}
        \caption{Camera positions and directions by Mapillary Images}
    \end{subfigure}
    
        \vspace{1em}
    
    \begin{subfigure}[b]{0.45\textwidth}
        \centering
        \includegraphics[width=\textwidth]{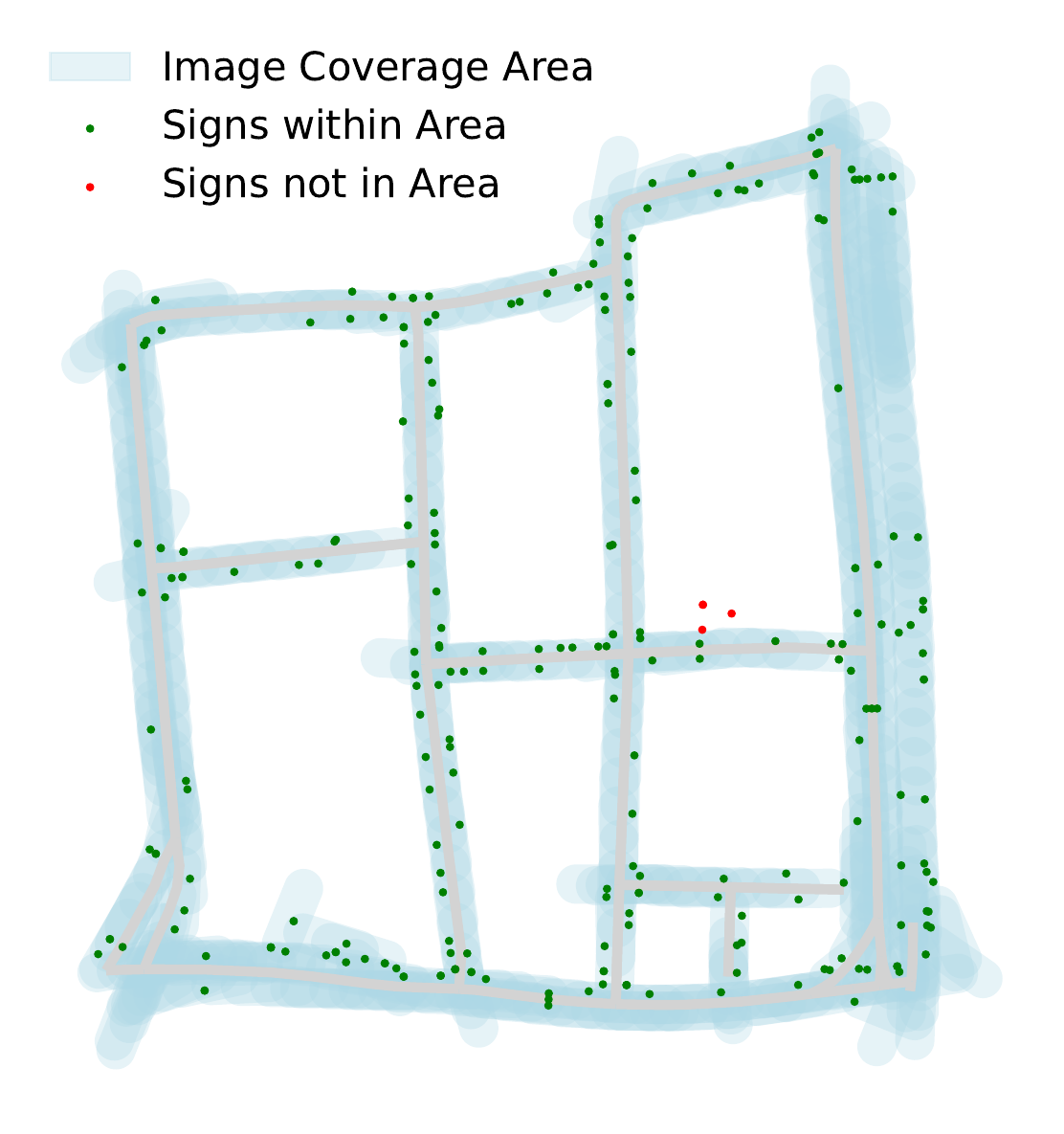}
        \caption{Theoretically visible signs on Cyclomedia Images}
    \end{subfigure}
    \hfill
    \begin{subfigure}[b]{0.45\textwidth}
        \centering
        \includegraphics[width=\textwidth]{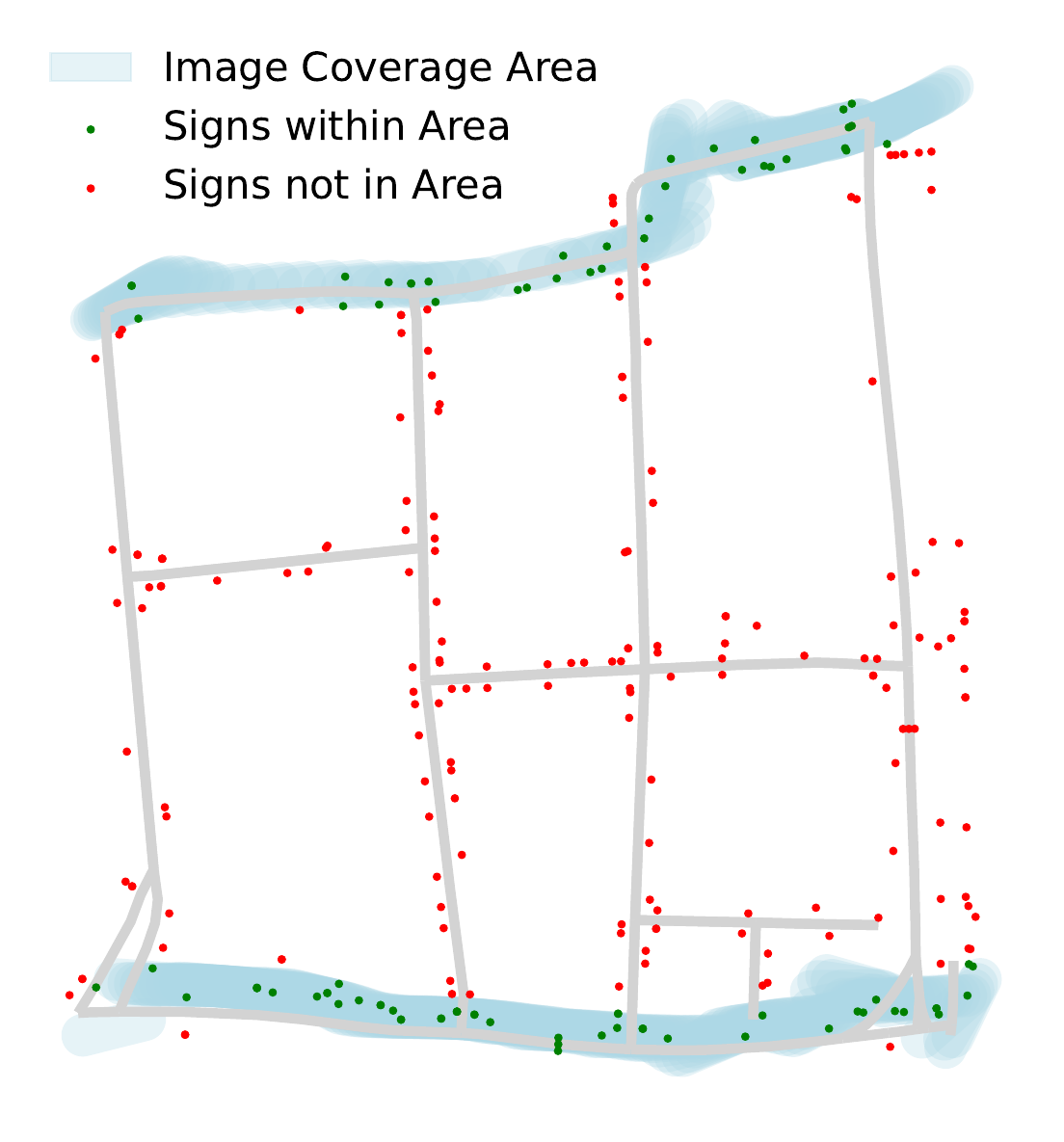}
        \caption{Theoretically visible signs on Mapillary Images}
    \end{subfigure}

    \caption{Positions and viewing directions of all recordings from both image sources (a, b) and all signs that should be visible within this covered area (c, d). Best viewed in colors.}
    \label{fig:image_coverage2}
\end{figure*}
\clearpage
\twocolumn
\section{Traffic Signs: Segmentation}

We chose three models to segment the traffic signs from the images
for comparison:
\begin{enumerate}
    \item a U-Net model trained on the A2D2 dataset,
    \item a SegFormer model fine-tuned on the A2D2 dataset,
    \item a Mask2Former model trained on the Mapillary Vistas dataset.
\end{enumerate}
We annotated 100 images of our Cyclomedia dataset and calculated the Intersection over Union (IoU) between the annotated and the predicted masks for each image. The average over all annotated images is shown in Table \ref{tab:seg_signs}. The models trained on the A2D2 dataset perform significantly better.
We assume the signs look more similar, as they were also recorded in Germany, just as in our test images. Based on those results, we decided to use the U-Net model in further steps.
Upon manual inspection, we noted that signs appearing large in the images are usually better segmented than small signs in the background.

\begin{table}[ht]
    \centering
    \begin{tabular}{l@{\hspace{5pt}}c@{\hspace{5pt}}c@{\hspace{5pt}}}
        \toprule
        Model       & Training Dataset  & Mean IoU \\
        \midrule
        U-Net       & A2D2              & 0.550 \\
        Segformer   & A2D2              & 0.479 \\
        Mask2Former & Mapillary Vistas  & 0.351 \\
        \bottomrule
    \end{tabular}
    \caption{Average IoU scores using different segmentation models for traffic signs. 100 images from Leipzig were manually annotated for this comparison.}
    \label{tab:seg_signs}
\end{table}

\clearpage
\onecolumn
\section{Traffic Signs: Deviation Box Plots}
\begin{figure}[!ht]
    \centering
    \begin{subfigure}[b]{0.45\textwidth}
        \centering
        \includegraphics[width=\textwidth]{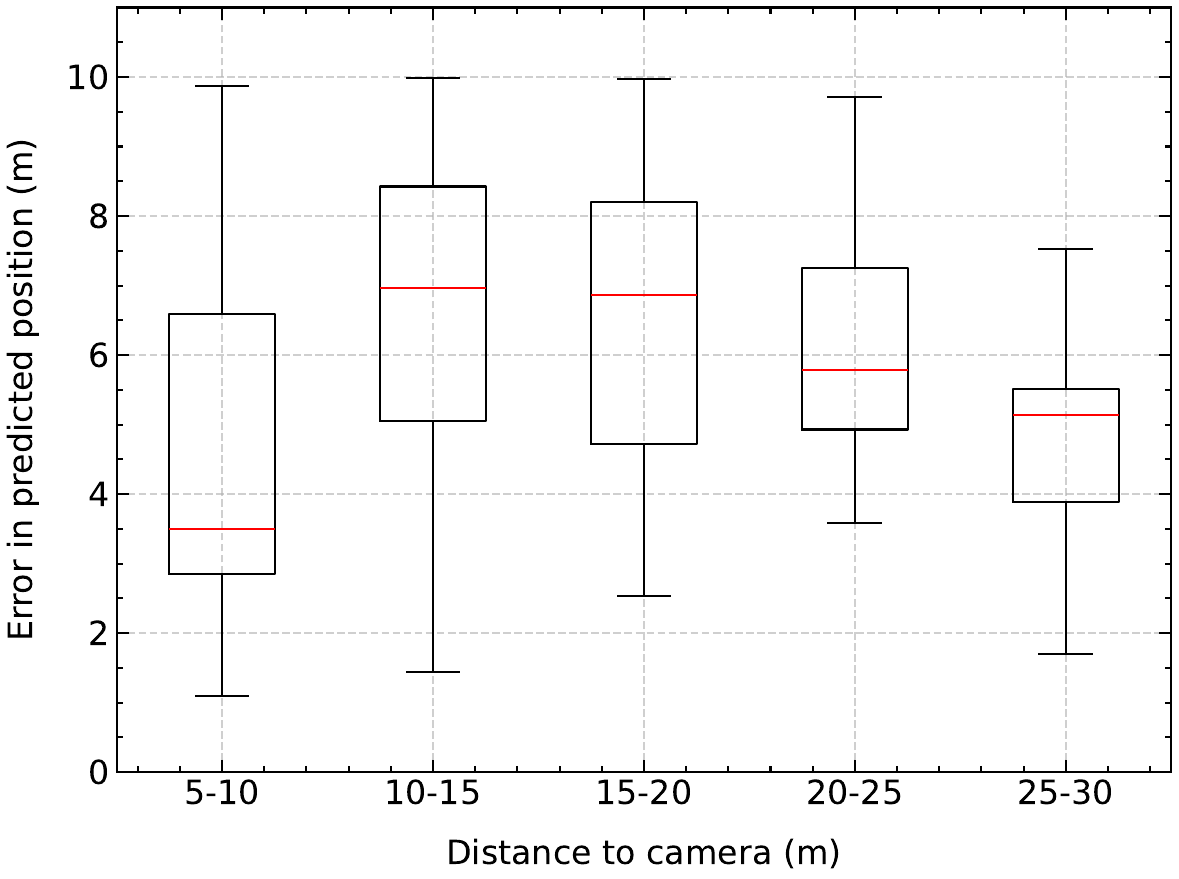}
        \caption{DepthAnything-B}
    \end{subfigure}
    \hfill
    \begin{subfigure}[b]{0.45\textwidth}
        \centering
        \includegraphics[width=\textwidth]{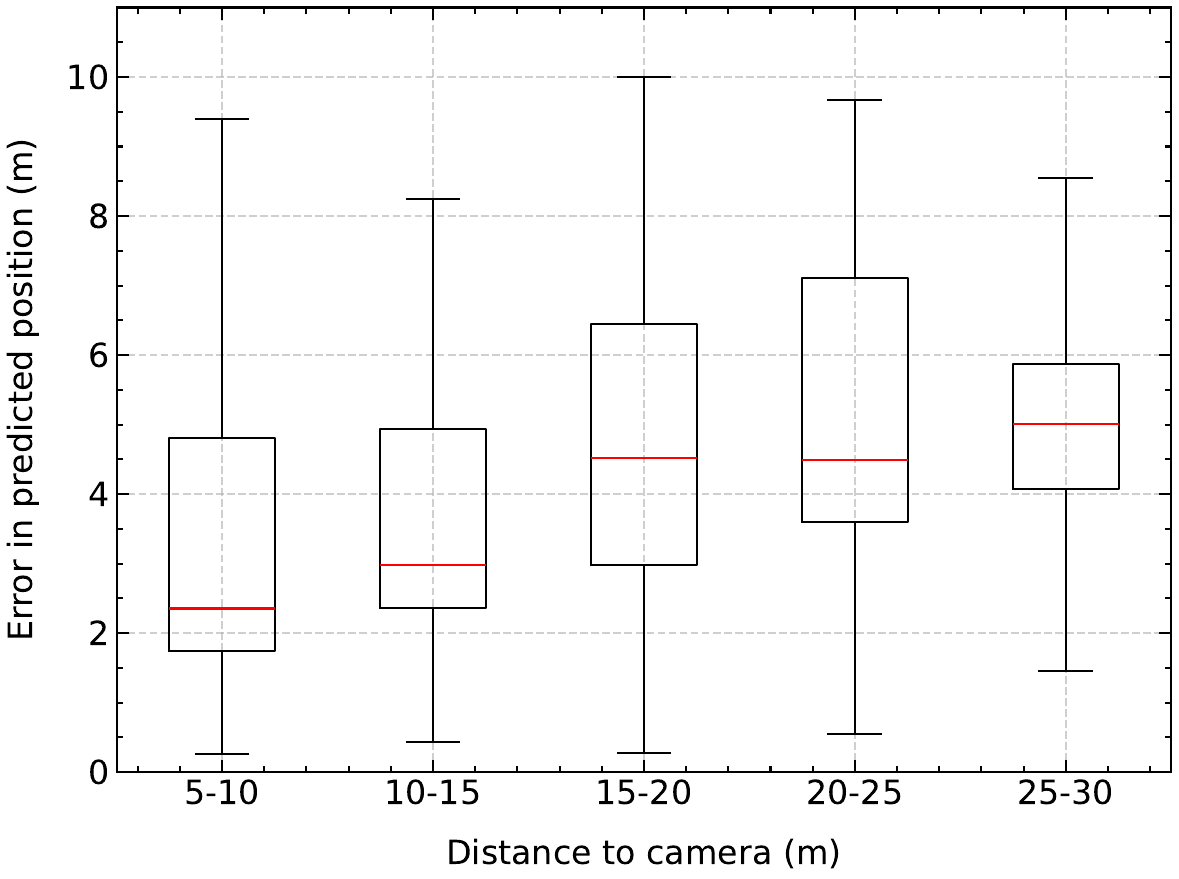}
        \caption{DepthPro (with cam)}
    \end{subfigure}
    
    \vspace{1em}
    
    \begin{subfigure}[b]{0.45\textwidth}
        \centering
        \includegraphics[width=\textwidth]{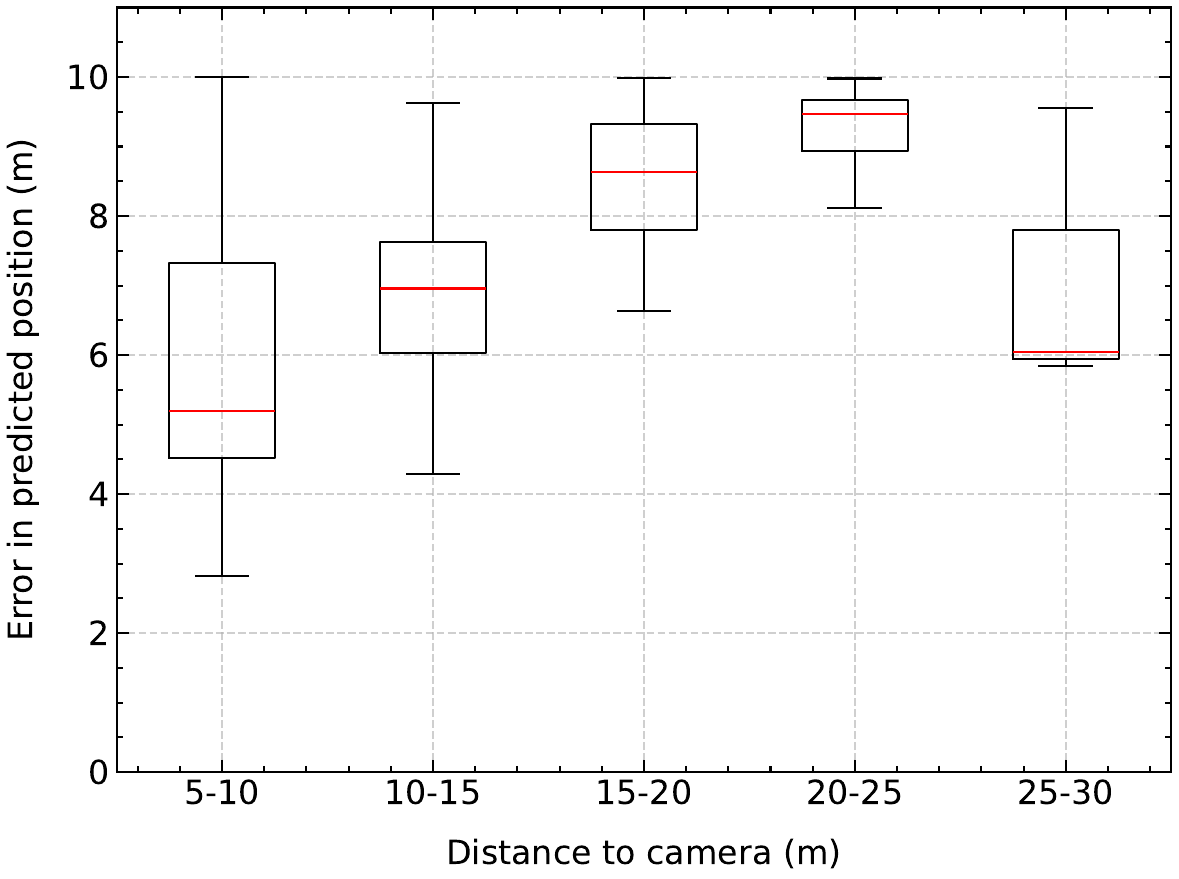}
        \caption{Metric3D-ViT}
    \end{subfigure}
    \hfill
    \begin{subfigure}[b]{0.45\textwidth}
        \centering
        \includegraphics[width=\textwidth]{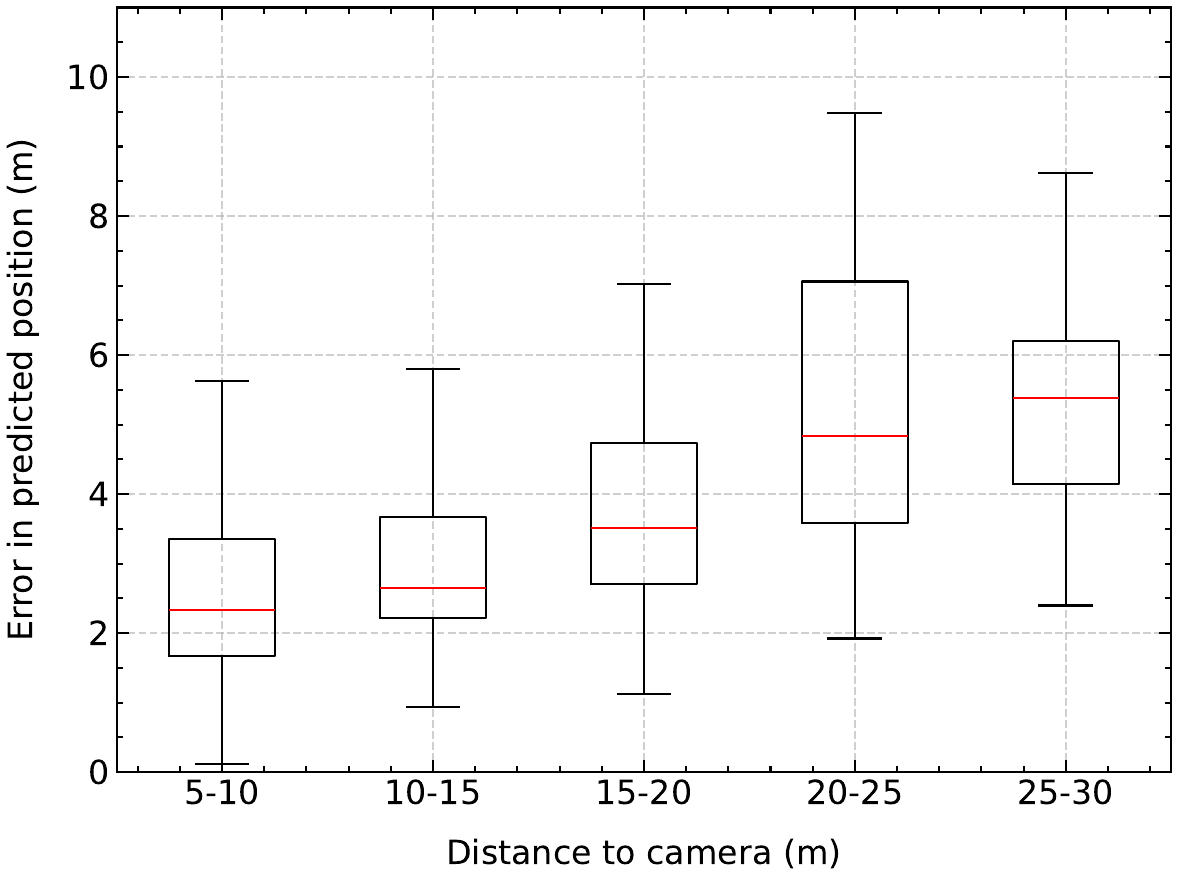}
        \caption{UniDepth (no cam)}
    \end{subfigure}

    \caption{The box plots show the relationship between the error of the position (y-axis) and the true distance between sign and camera (x-axis) for the signs that were matched with an annotation entry.}
    \label{fig:signs_box_plots}
\end{figure}

\begin{multicols}{2}
The box plots in Fig. \ref{fig:signs_box_plots} display the error in predicted position (y-axis) as a function of distance to the camera (x-axis) for four depth estimation methods. The plots provide insights into the accuracy and variability of predictions across distance intervals.
The first plot (DepthAnything-B) shows a non-monotonic trend: errors start low, peak in the mid-range (~7m at 15-20m), and decrease at greater distances (~4.5m at 25-30m).
The second plot (DepthPro) shows a consistent increase in median error with distance increasing, rising steadily from ~2.3m (5-10m) to ~5m (25-30m). Variability grows predictably with distance, reflecting stable performance.
The third plot (Metric3D) shows higher overall errors, with a median of ~9m even at close ranges. Errors fluctuate with distance, peaking at 20-25m (median 9.5m) before slightly decreasing at 25-30m. This irregular trend suggests challenges in specific ranges.
The fourth plot (UniDepth) resembles the UniDepth plot, with errors starting at 2.3m for 5-10m and reaching 5.3m at 25-30m. The interquartile range (IQR) and whisker lengths also grow with distance, indicating increasing variability.
Overall, UniDepth and DepthPro exhibit predictable behavior, whereas Metric3D and DepthAnything exhibit irregularities.
\end{multicols}
\clearpage
\onecolumn
\section{Road Damages: Deviation Box Plots}

\begin{figure}[ht!]
    \centering
    \begin{subfigure}[b]{0.45\textwidth}
        \includegraphics[width=\textwidth]{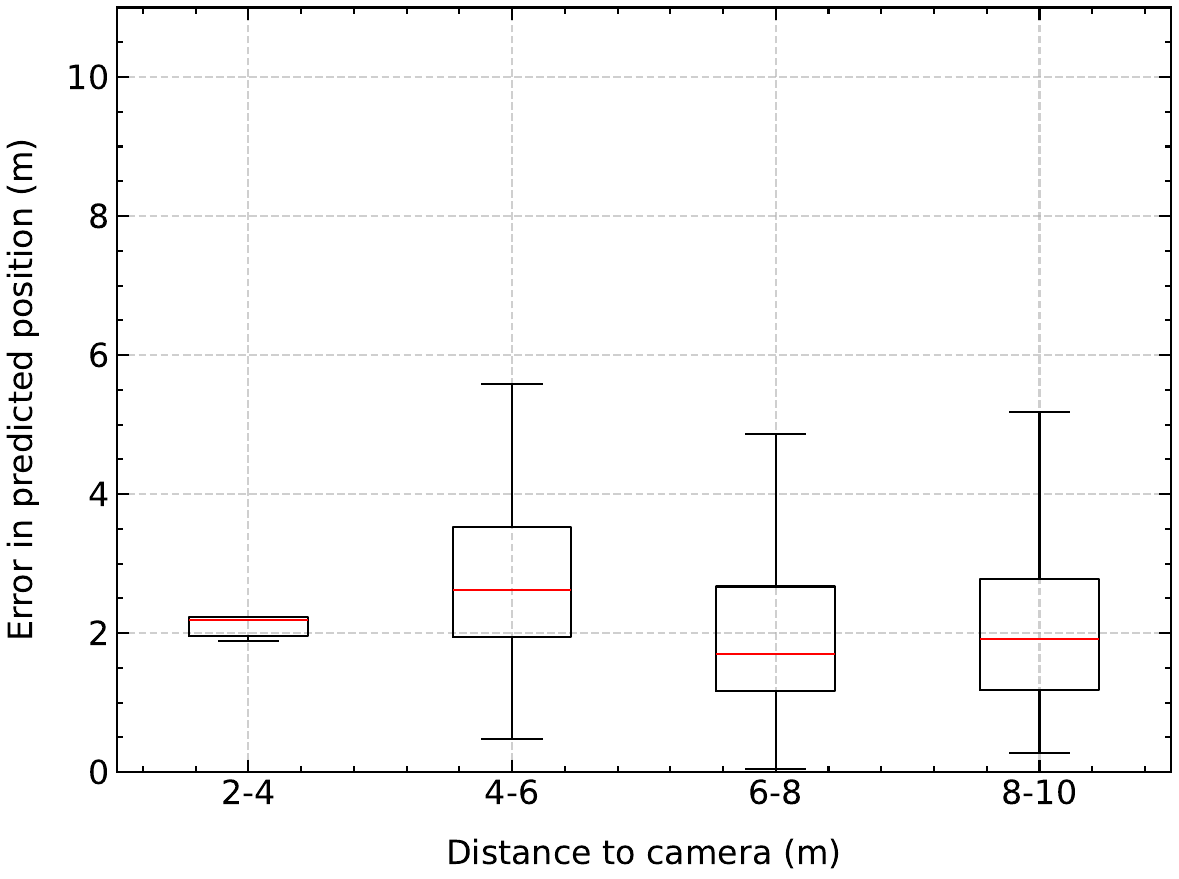}
        \caption{DepthAnything-B}
    \end{subfigure}
    \hfill
    \begin{subfigure}[b]{0.45\textwidth}
        \includegraphics[width=\textwidth]{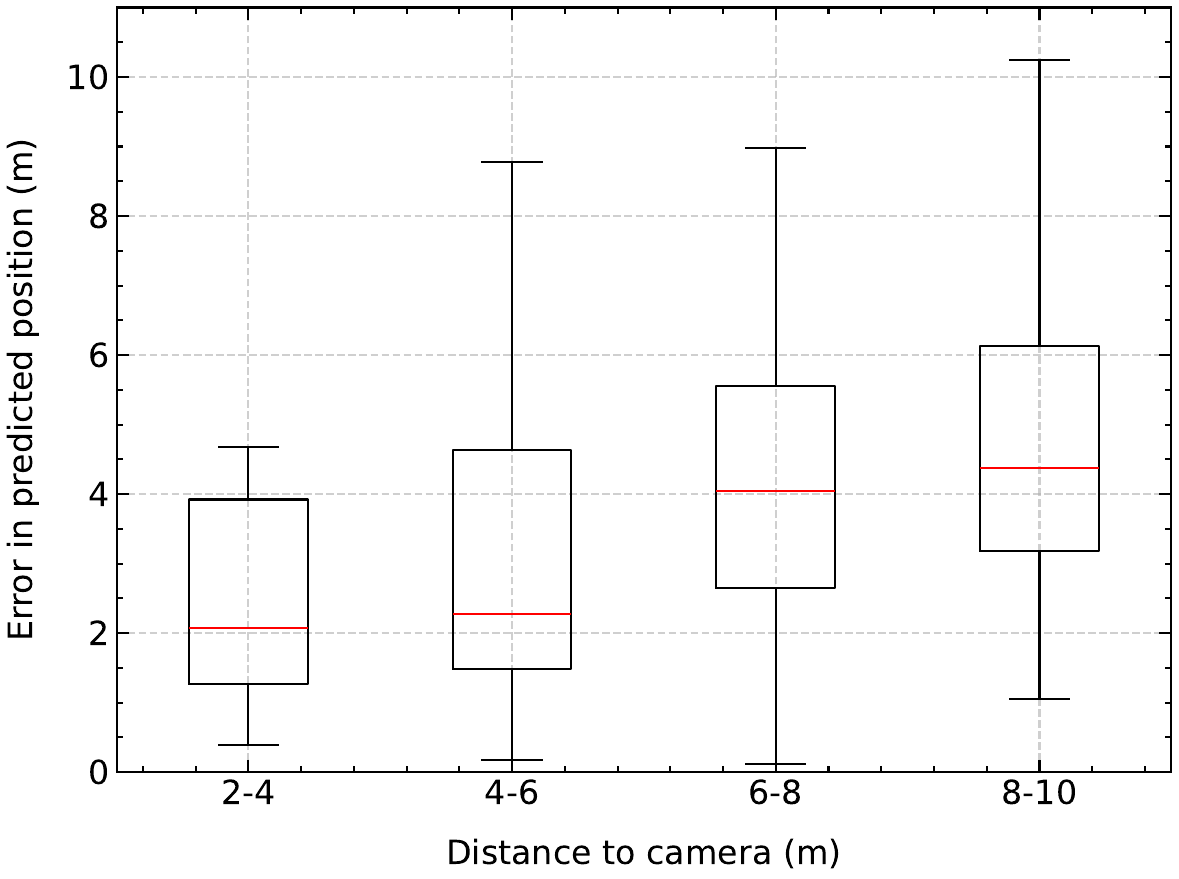}
        \caption{DepthPro (with cam)}
    \end{subfigure}

    \vspace{1em}

    \begin{subfigure}[b]{0.45\textwidth}
        \includegraphics[width=\textwidth]{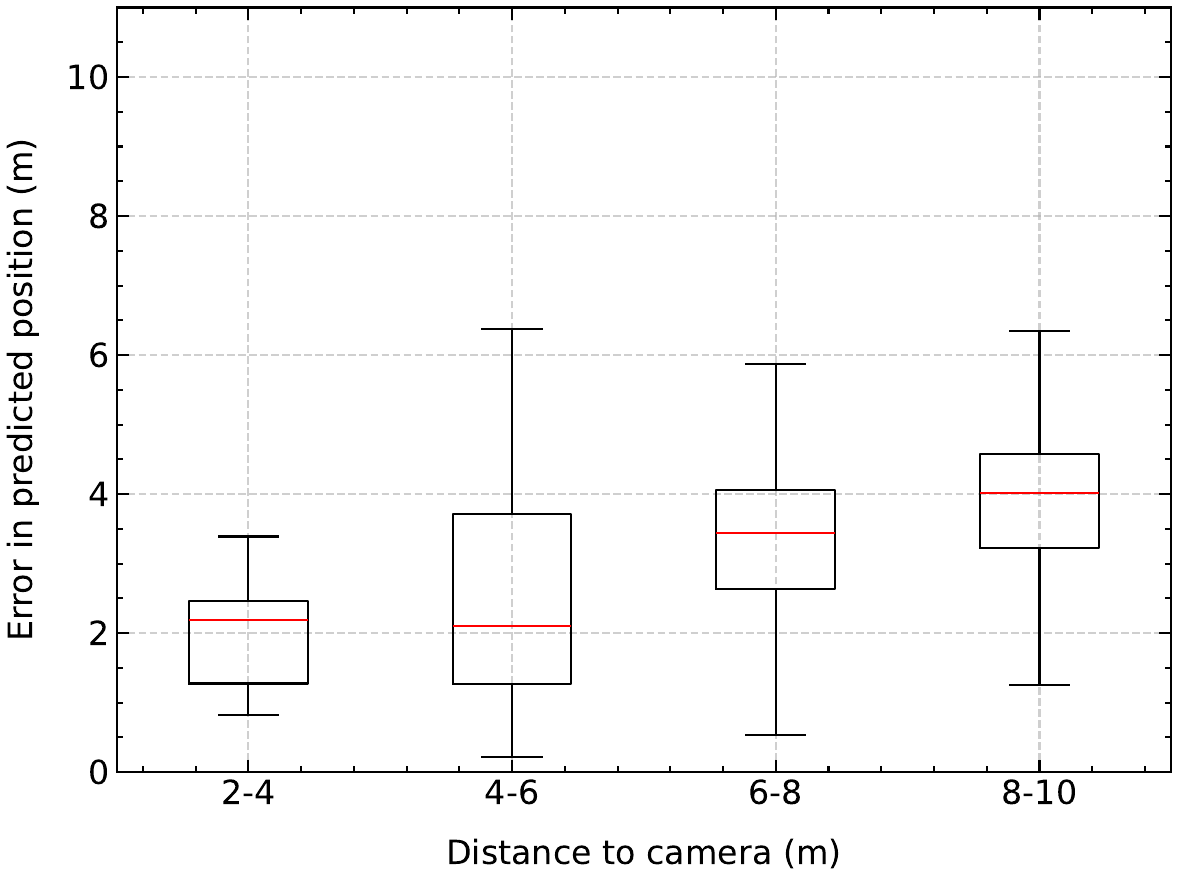}
        \caption{Metric3D-ViT}
    \end{subfigure}
    \hfill
    \begin{subfigure}[b]{0.45\textwidth}
        \includegraphics[width=\textwidth]{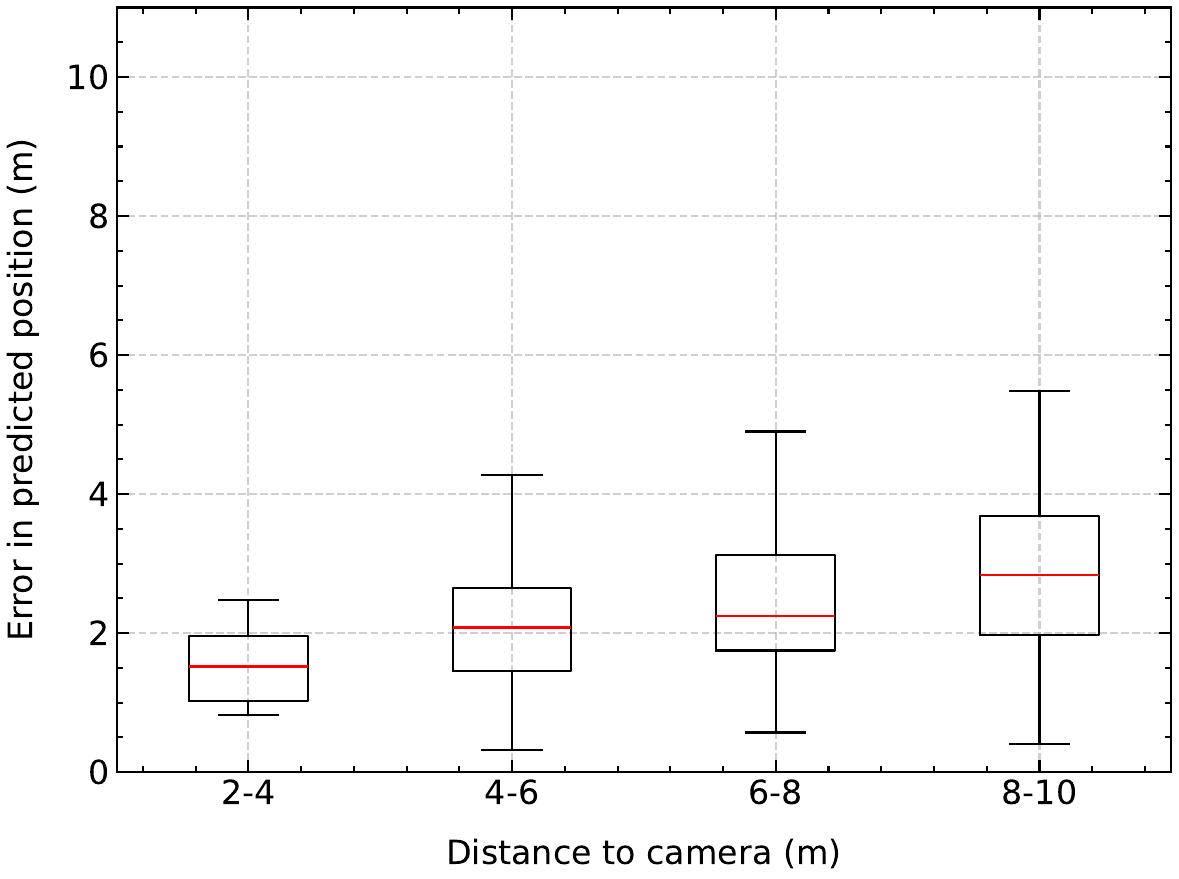}
        \caption{UniDepth (no cam)}
    \end{subfigure}

    \caption{The box plots show the relationship between the error of the position (y-axis) and the true distance between road damage and camera (x-axis).}
    \label{fig:damages_box_plots}
\end{figure}

\twocolumn

\end{document}